\documentclass[letterpaper, 10 pt, conference]{ieeeconf}  
\IEEEoverridecommandlockouts                              

\makeatletter
\let\NAT@parse\undefined
\makeatother

\usepackage{graphics} 
\usepackage{subcaption}
\usepackage{graphicx}
\usepackage{grffile} 
\usepackage{amsmath} 
\usepackage{amssymb}  
\usepackage{color}
\usepackage{cite}
\usepackage{bm}
\usepackage{url}

\usepackage[noend]{algpseudocode}
\usepackage{algorithm}
\usepackage[table]{xcolor}
\usepackage{makecell}
\usepackage{multirow}
\usepackage[draft]{todonotes}
\usepackage[colorlinks,bookmarks=false]{hyperref}

\def\etal{et al.}
\definecolor{color_no}{rgb}{1.        ,  1.        ,  1.        }
\definecolor{color_ed}{rgb}{0.        ,  0.15882353,  1.        }
\definecolor{color_pl}{rgb}{0.        ,  0.83333333,  1.        }
\definecolor{color_sp}{rgb}{0.49019608,  1.        ,  0.47754586}
\definecolor{color_cy}{rgb}{1.        ,  0.90123457,  0.        }
\definecolor{color_co}{rgb}{1.        ,  0.27668845,  0.        }
\definecolor{color_nu}{rgb}{0.5       ,  0.        ,  0.        }

\title{\LARGE \bf
Primitive Fitting Using Deep Boundary Aware Geometric Segmentation
}

\author{Duanshun Li$^{1*}$ and Chen Feng$^{2*}$
\thanks{*This work was supported by Mitsubishi Electric Research Labs (MERL) and New York University. The authors contributed equally.}
\thanks{$^{1}$Duanshun Li is with University of Alberta, 116 St \& 85 Ave, Edmonton, AB T6G 2R3, Canada,
        {\tt\small duanshun@ualberta.ca}}%
\thanks{$^{2}$Chen Feng is the corresponding author and is with New York University, 6 MetroTech Center,
	Brooklyn, NY 11201, USA
	{\tt\small cfeng@nyu.edu}}%
}

\begin{document}

\bstctlcite{IEEEexample:BSTcontrol}

\maketitle
\thispagestyle{empty}
\pagestyle{empty}

\begin{abstract}

To identify and fit geometric primitives (e.g., planes, spheres, cylinders, cones) in a noisy point cloud is a challenging yet beneficial task for fields such as robotics and reverse engineering. As a multi-model multi-instance fitting problem, it has been tackled with different approaches including RANSAC, which however often fit inferior models in practice with noisy inputs of cluttered scenes. Inspired by the corresponding human recognition process, and benefiting from the recent advancements in image semantic segmentation using deep neural networks, we propose BAGSFit as a new framework addressing this problem. Firstly, through a fully convolutional neural network, the input point cloud is point-wisely segmented into multiple classes divided by jointly detected instance boundaries without any geometric fitting. Thus, segments can serve as primitive hypotheses with a probability estimation of associating primitive classes. Finally, all hypotheses are sent through a geometric verification to correct any misclassification by fitting primitives respectively. We performed training using simulated range images and tested it with both simulated and real-world point clouds. Quantitative and qualitative experiments demonstrated the superiority of BAGSFit.

\end{abstract}

\section{Introduction}

\begin{quote}
	\textit{``Treat nature by means of the cylinder, the sphere, and the cone.'' \hfill --- Paul C{\'e}zanne, 1904}
\end{quote}

Not only in art, the idea of decomposing a scene or a complex object into a set of simple geometric primitives for visual object recognition dates back as early as 1980s when Biederman proposed the object Recognition-By-Components theory~\cite{biederman1987recognition}, in which primitives were termed ``geons''.
Although some real scenes can be more complicated than simple combinations of ``geons'', 
there are many useful ones that can be efficiently modeled for the purpose of robotics:
planes in man-made structures,
utility pipelines as cylinders,
household objects such as paper cups,
and more interestingly, a robot itself, often as an assembly of simple primitives.

Thus, for better extro- and intro-spection to improve the intelligence of all kinds of robots,
from autonomous cars to service robots,
it is beneficial to robustly detect those primitives and 
accurately estimate the associated parameters from noisy 3D sensor inputs,
such as
robotic manipulation that requires poses and shapes of objects~\cite{berner2013combining},
SLAM that takes advantage of primitives (mostly planes) for better mapping accuracy~\cite{taguchi2013point,ma2016cpa,dzitsiuk2017noising},
reverse engineering that models complex mechanical parts as primitives~\cite{li2011globfit},
and similarly as-built Building Information Modeling~\cite{tang2010automatic,xiao2014reconstructing}.

\begin{figure} 
	\centering
	
	\frame{\includegraphics[width=.47\columnwidth]{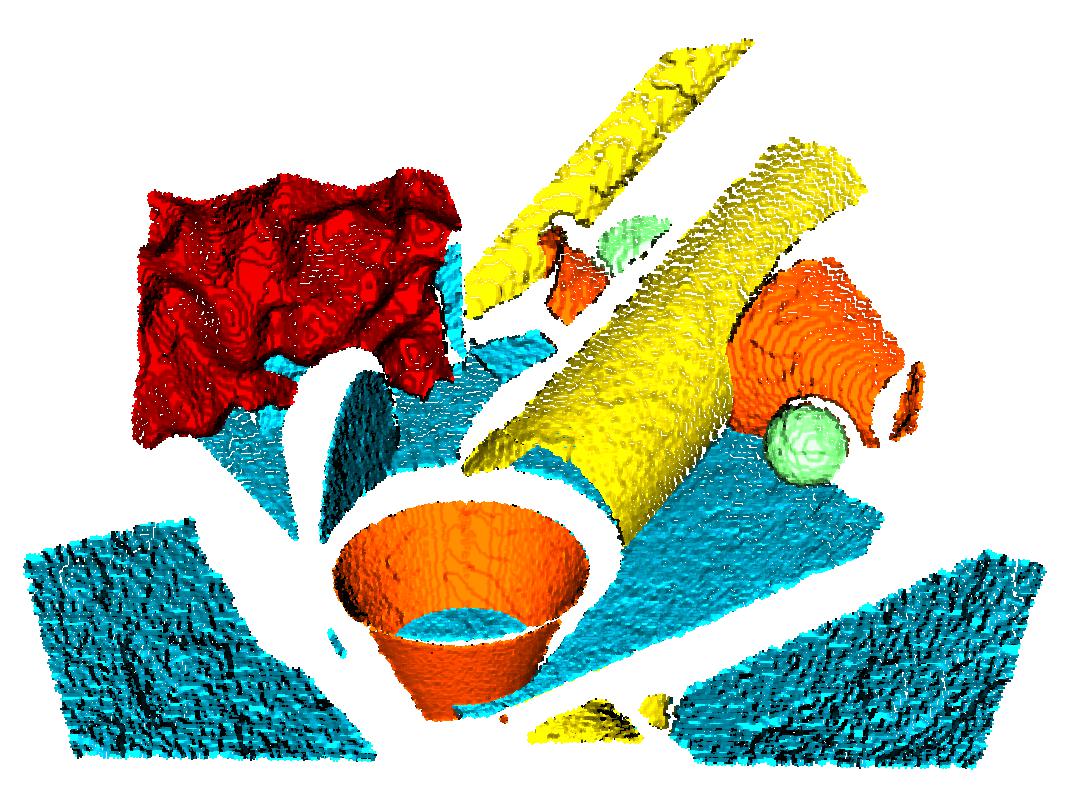}}
	\frame{\includegraphics[width=.47\columnwidth]{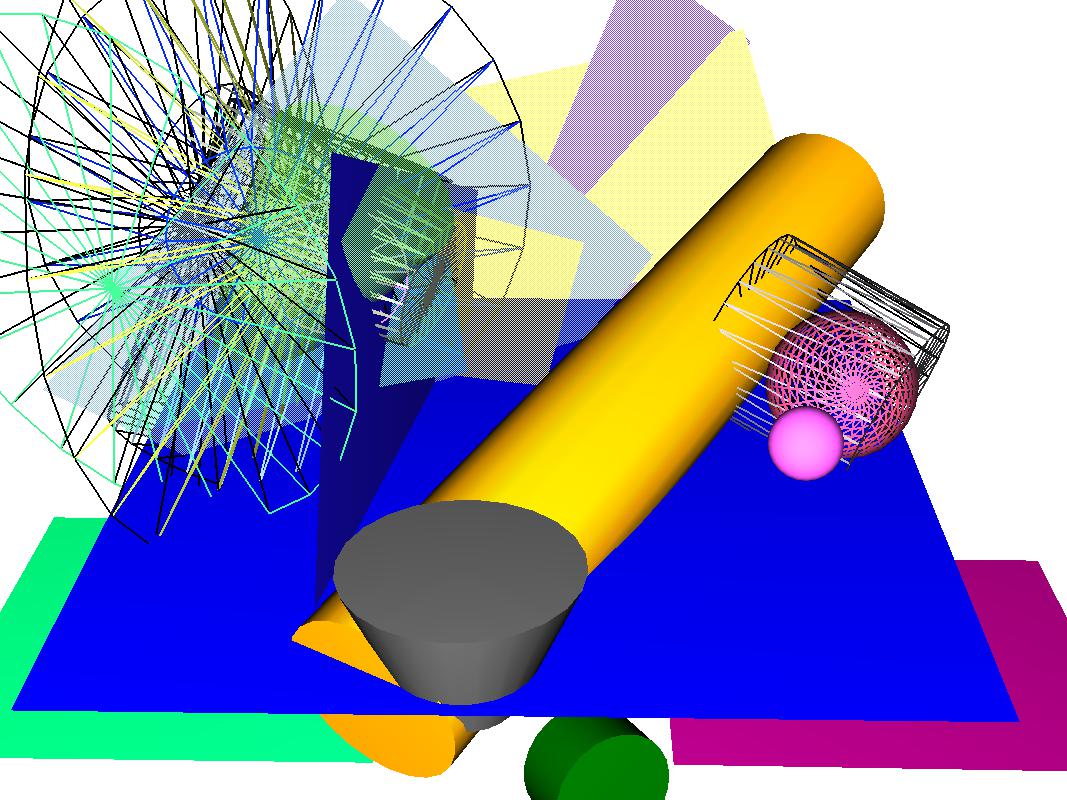}}
	\frame{\includegraphics[width=.47\columnwidth]{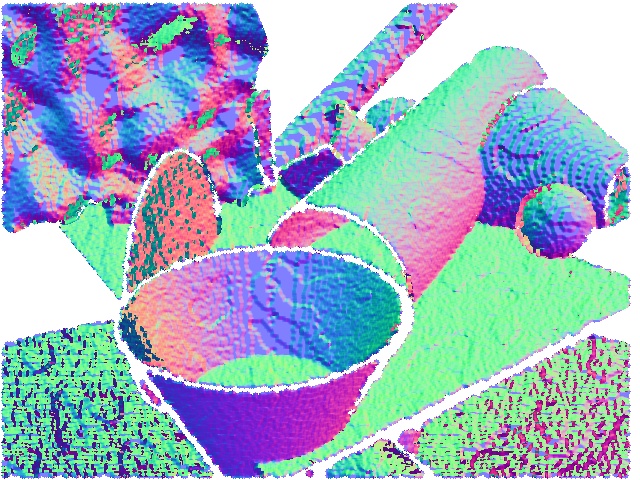}}
	\frame{\includegraphics[width=.47\columnwidth]{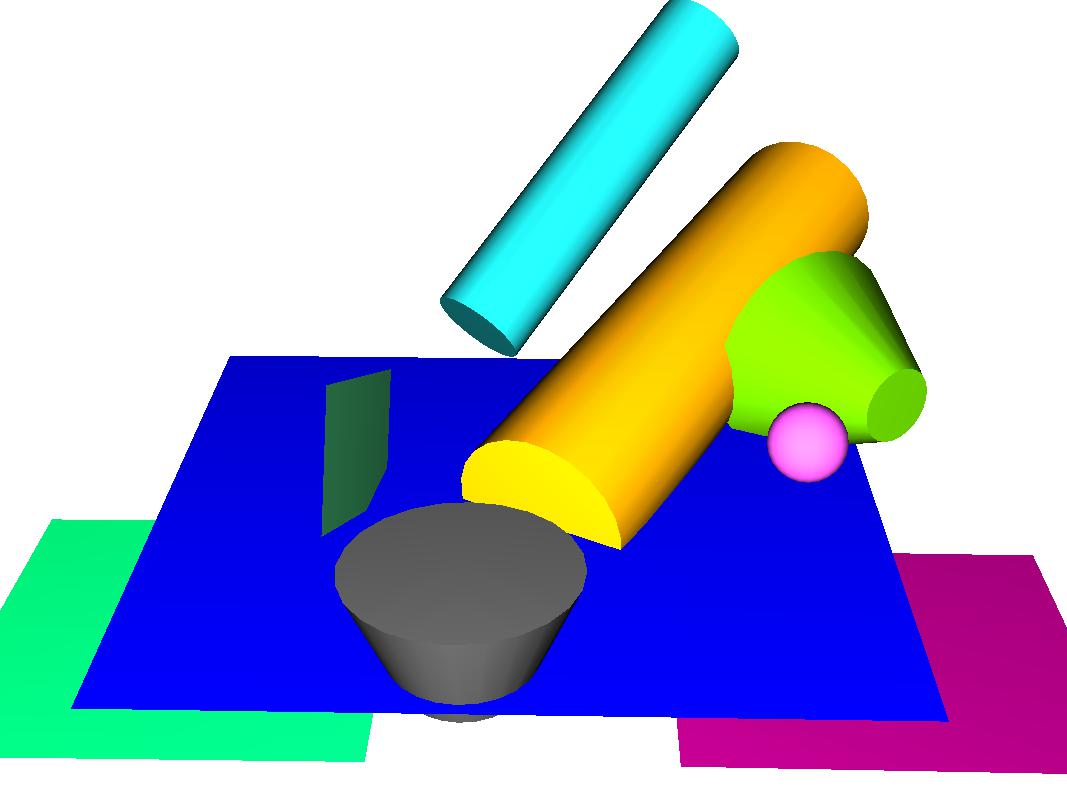}}
	\frame{\includegraphics[width=.47\columnwidth]{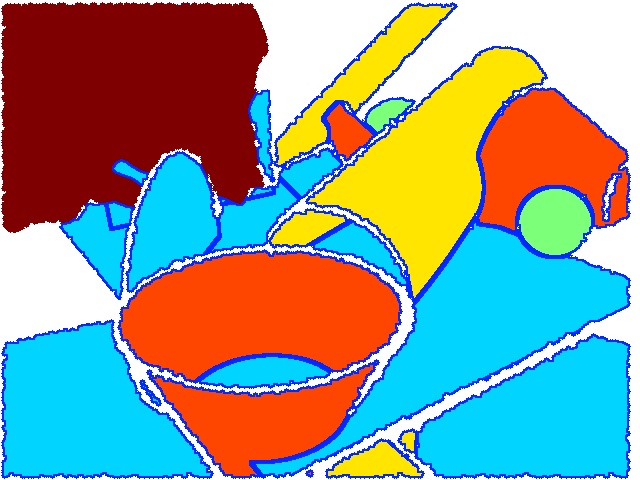}}
	\frame{\includegraphics[width=.47\columnwidth]{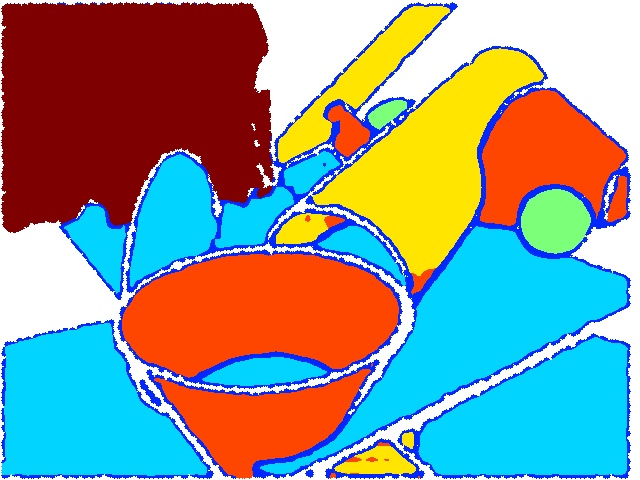}}
	
	\resizebox{0.94\columnwidth}{!}{
		\tiny
		\begin{tabular}{@{}cccccc@{}}
			\cellcolor{color_ed} \textcolor{white}{Boundary} &
			\cellcolor{color_pl} Plane &
			\cellcolor{color_sp} Sphere &
			\cellcolor{color_cy} Cylinder &
			\cellcolor{color_co} Cone &
			\cellcolor{color_nu} \textcolor{white}{Other}
		\end{tabular}
	}
	\caption{Primitive fitting on a simulated test range image (top left) with BAGSFit (middle right) vs. RANSAC (top right)~\cite{schnabel2007efficient}. Estimated normals (middle left) and ground truth labels (bottom left) are used to train a fully convolutional segmentation network in BAGSFit. During testing, a boundary-aware and thus instance-aware segmentation (bottom right) is predicted, and sent through a geometric verification to fit final primitives (randomly colored). Comparing with BAGSFit, the RANSAC-based method produces more misses and false detections of primitives (shown as transparent or wire-frame), and thus a less appealing visual result.
	\label{fig:teaser}}
\end{figure}

This primitive fitting is a classic chicken-and-egg problem:
with given primitive parameters, point-to-primitive (P2P) membership can be determined by nearest P2P distance;
and vise versa by robust estimation.
The challenge comes when multiple factors present together:
a noisy point cloud (thus noisy normal estimation), 
a cluttered scene due to multiple instances of a same or multiple primitive models,
and also background points not explained by the primitive library.
See Figure~\ref{fig:teaser} for an example.
The seminal RANSAC-based method~\cite{schnabel2007efficient} often tends to fit inferior primitives
that do not well represent the real scene.

Different from existing work for this multi-model multi-instance fitting problem,
we are inspired by human visual perception of 3D primitives.
As found by many cognitive science researchers,
human ``observers' judgments about 3D shape are often systematically distorted''~\cite{todd2004visual}.
For example, when looking at a used fitness ball, many people would think of it as a sphere,
although it could be largely distorted if carefully measured.
This suggests that human brain might not be performing exact geometric fitting during primitive recognition,
but rather rely on ``qualitative aspects of 3D structure''~\cite{todd2004visual} from visual perception.

Due to the recent advancements in image semantic segmentation using convolutional neural networks (CNN)~\cite{chen2016deeplab,yu2017casenet},
it is then natural to ask: whether CNN can be applied to this problem with geometric nature
and find the P2P membership as segmentation on the range image without geometric fitting at all?
Our answer is yes, which leads to the BAGSFit framework that reflects this thought process.

\subsection{Contributions}

This paper contains the following key contributions:

\begin{itemize}
	\item We present a methodology to easily obtain point-wise ground truth labels from simulated dataset for supervised geometric segmentation, demonstrate its ability to generalize to real-world dataset, and released the first simulated dataset \footnote{The dataset is publicly
		available at \url{https://github.com/ai4ce/BAGSFit}.} for development and benchmarking.
	\item We present a novel framework for multi-model 3D primitive fitting, which performs both qualitatively and quantitatively superior than RANSAC-based methods on noisy range images of cluttered scenes.
	\item We introduce this geometric segmentation task for {CNN} with several design analyses and comparisons.
\end{itemize}

\subsection{Related Work}

Depending on the basic assumption of the number of primitive classes, previous work can be roughly grouped into multi-instance vs. multi-model, described as follows.

\textbf{Multi-Instance Fitting}: As a subset of multi-model fitting problem,
it typically assumes that a scene is mainly composed of a single class of primitive model,
such as plane detection by normal-based region grow~\cite{holz2011real} or agglomerative clustering~\cite{feng2014fast},
and sphere detection by hough transform~\cite{abuzaina2013sphere} or mean-shift clustering~\cite{tran2016esphere}.
Benefiting from the simple assumption, such methods can work in real-time and obtain accurate results.
Yet when the assumption is violated, they often generate many false detections (e.g., a curved surface fitted as multiple small planes),
which requires careful threshold tuning to be filtered out.

\textbf{Multi-Model Fitting}: By assuming the existence of potentially multiple classes of primitives in a scene,
it is more realistic than the previous group, and thus more challenging when a cluttered scene is observed with noisy 3D sensors.
Previous work with this assumption can be roughly grouped further into the following three categories:

\textit{Segmentation}: These methods are rooted in the idea of segmentation of a point cloud into individual clusters,
while the model classification and fitting are either performed during the segmentation or afterwards.
For example, an early method simultaneous grows multiple seed regions with dynamic primitive model selection by iterative regression~\cite{leonardis1990segmentation}, but was not shown to necessarily work for noisy cluttered scenes.
A more recent work in reverse engineering~\cite{toony2015describing} assumes the input 3D mesh has been previously segmented into parts,
and then classifies each part based on the Gaussian sphere, i.e. the normal space.
The data quality in this work is much better than that of the more prevalent Kinect-like range images,
thus might not be suitable for noisy and cluttered data in robotics.
Another recent real-time algorithm~\cite{georgiev2016real} fits conic curves in each scan-line and then merges neighboring curves into primitives.
Maybe due to miss detections of smaller curves, this first 2D then 3D approach has not been shown to work robustly in cluttered scene.
While sharing the same basic segmentation idea as above, our geometric segmentation CNN performs no fitting,
but only provides plausibility maps for further geometric verification if needed, or even just as shape priors to more complex scene reconstruction~\cite{martens2017geometric}, e.g., tree trunks as cylindrical shapes.

\textit{RANSAC}: Since the seminal work by Schnabel \etal~\cite{schnabel2007efficient},
the idea of selecting sampled primitive hypotheses to maximize some scoring functions becomes a default solution to this problem,
serving as our baseline method.
It stimulates several variations including the {GlobFit} exploring spatial constraints between primitives for regularization~\cite{li2011globfit}.
Note in primitive fitting practice, the 3D sensor noise is often more structured (e.g., depth dependent noises for range images) than uniform or Gaussian in 3D as experimented in many of these papers.
What really makes the problem difficult is that those noisy points belonging to other partially occluded primitive instances become outliers of the primitive to be fit at hand,
causing false detections of ``ghost'' primitives not existed in the real scene but still with very small fitting errors and large consensus scores,
e.g. the ghost cones fitted with cylinder and background points in the top right of Figure~\ref{fig:teaser}.
More recently, prior probabilities or quality measure of the data~\cite{Chum2005PROSAC,TordoffMLESAC} were used to improve the probability of sampling an all-inlier subset.
Others explored the use of spatial consistency between the data~\cite{Sattler2009SCRAMSAC,Ni2009GroupSAC} to speed-up the hypothesis generation process. 
RANSAC has also been generalized as a set coverage problem~\cite{magri2016multiple}
or extended from a duality perspective as preference analysis or residual sorting and its variants~\cite{toldo2008robust,chin2012ransac,pham2014random,magri2015robust}.
While being theoretically interesting with good performances on classic multi-instance 2D fitting tasks such multi-homography detection, we are not aware of any of those methods that are explicitly shown to work well in the setting of multi-modal geometric primitive fitting from cluttered and noisy 3D range images.

\textit{Energy Minimization}: Unlike the sequential and greedy nature of RANSAC based methods, it is appealing in theory to define a global energy function in terms of P2P membership that once minimized results in desired solution~\cite{isack2012energy,woodford2012contraction,barath2017multi,amayo2017geometric}.
However most of them are only shown on relatively small number of points of simple scenes without much clutters or occlusions,
and it is unclear how they will scale to larger datasets due to the intrinsic difficulty and slowness of minimizing the energy function.


\begin{figure*}[t!] 
	\centering
	\begin{subfigure}{0.4\textwidth}
		\includegraphics[width=.48\textwidth]{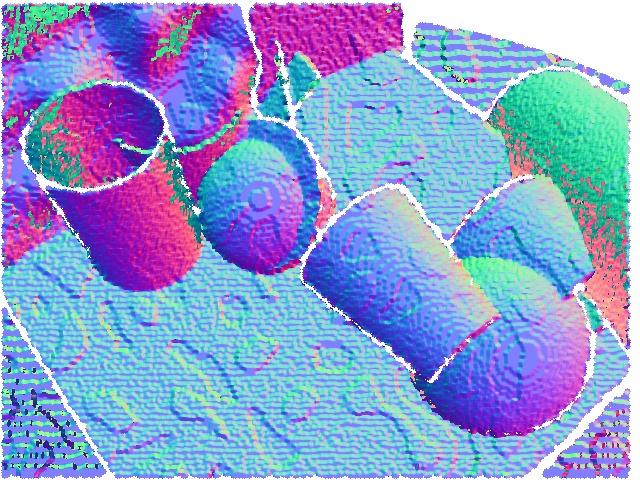}\hspace{0em}%
		\includegraphics[width=.5\textwidth]{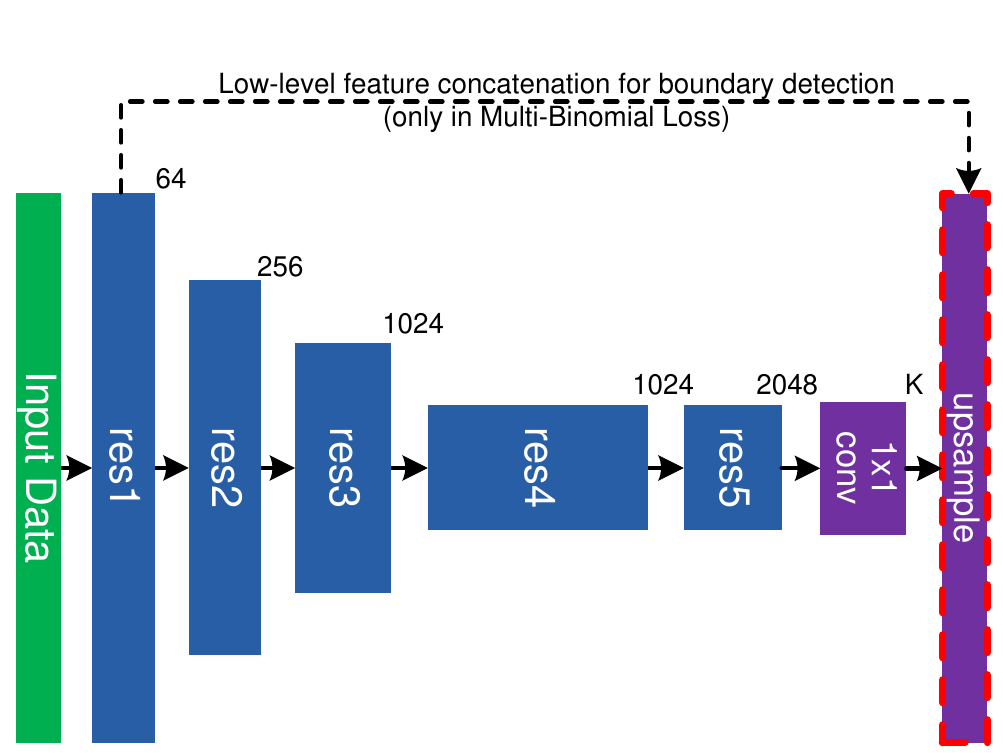}
		\caption{CNN-based Segmentation\label{fig:overview_seg}}
	\end{subfigure}\hspace{0em}%
	\begin{subfigure}{0.4\textwidth}
		\frame{\includegraphics[width=0.24\textwidth]{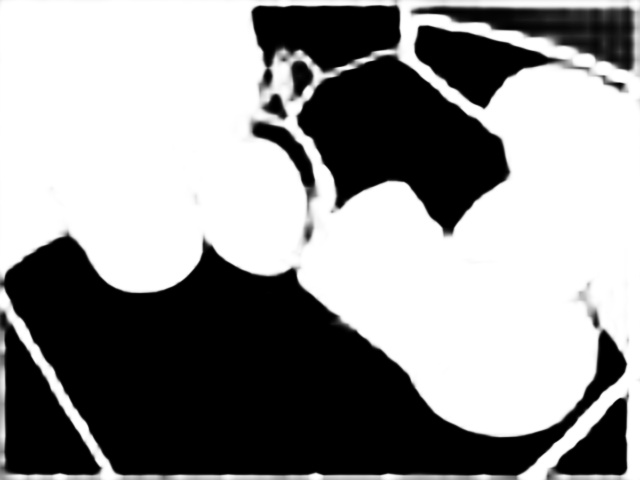}}\hspace{0em}%
		\frame{\includegraphics[width=0.24\textwidth]{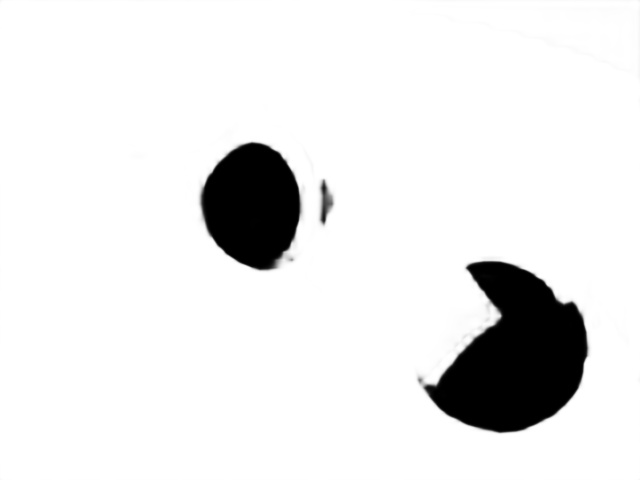}}\hspace{0em}%
		\frame{\includegraphics[width=0.24\textwidth]{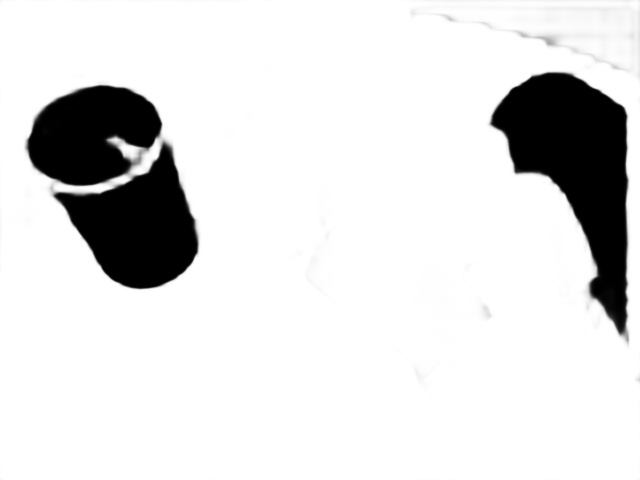}}\hspace{0em}%
		\frame{\includegraphics[width=0.24\textwidth]{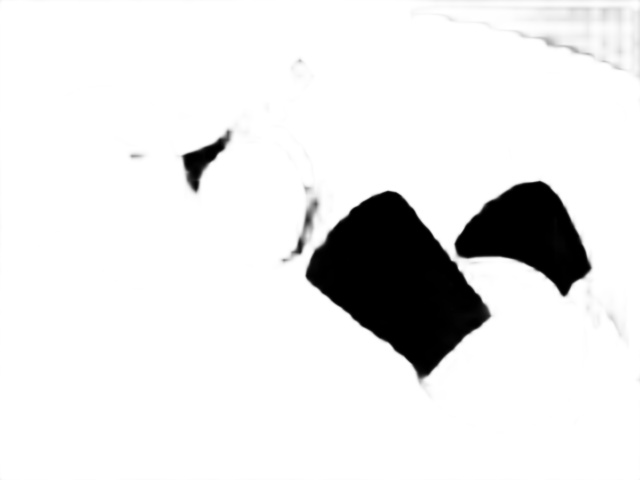}}\hspace{0em}%
		\frame{\includegraphics[width=0.24\textwidth]{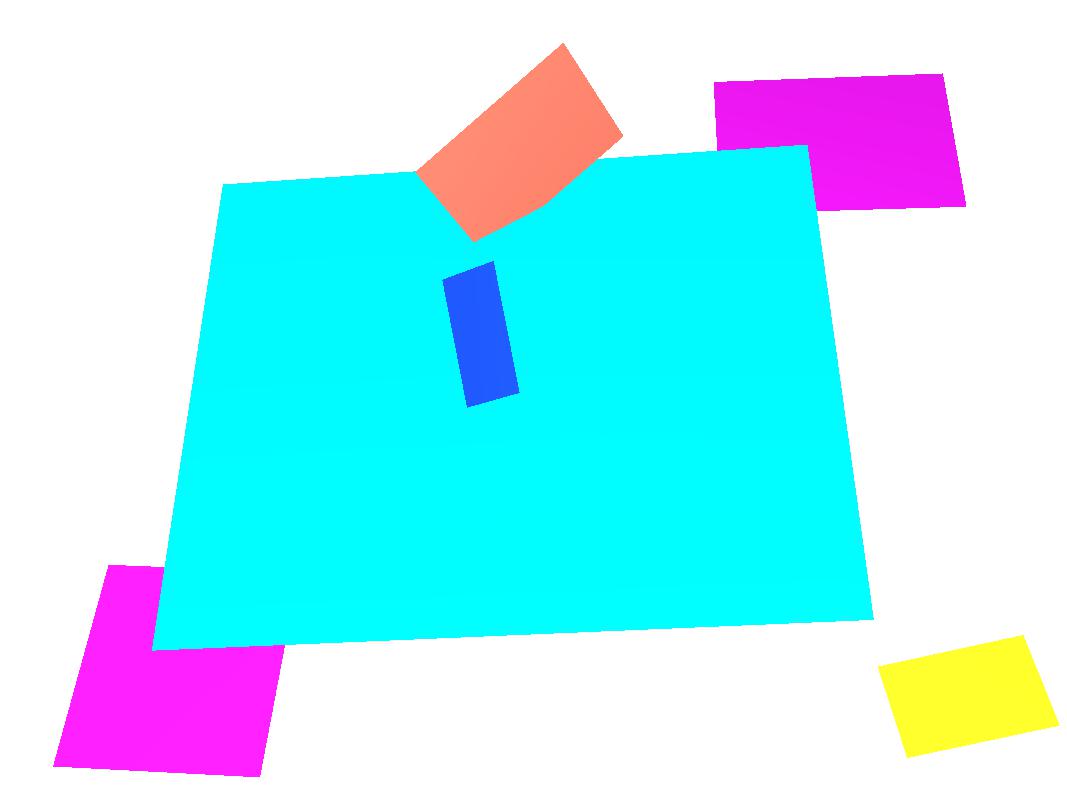}}\hspace{0em}%
		\frame{\includegraphics[width=0.24\textwidth]{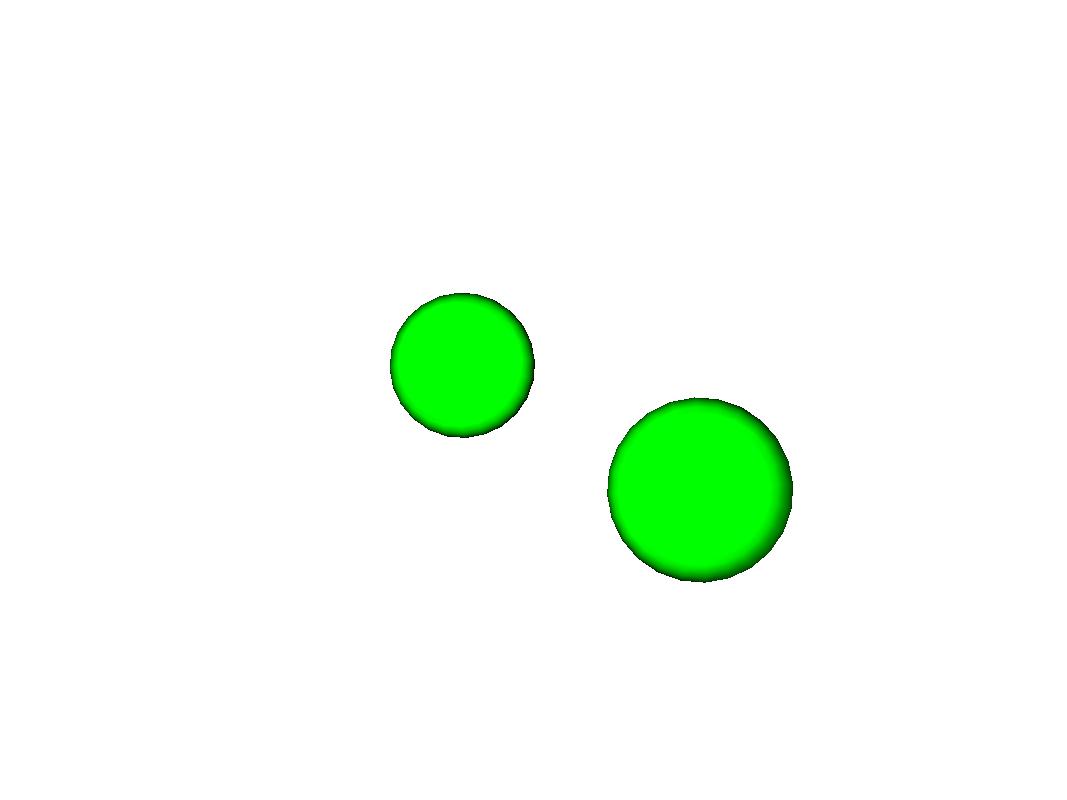}}\hspace{0em}%
		\frame{\includegraphics[width=0.24\textwidth]{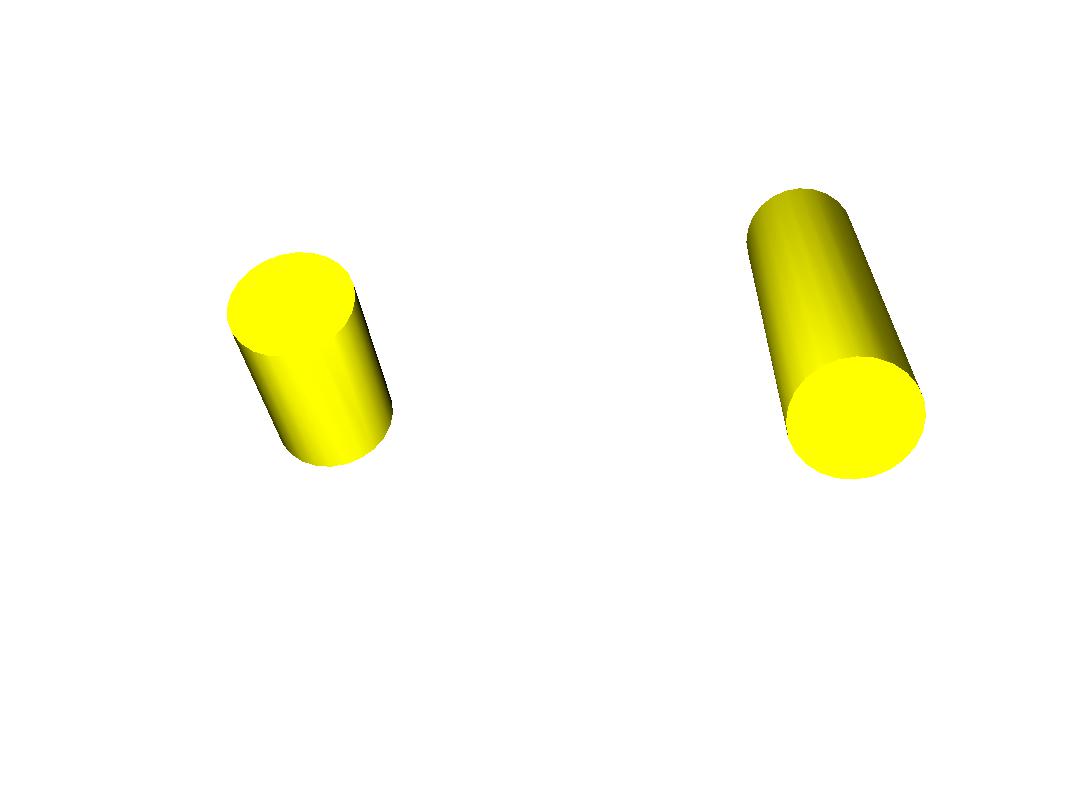}}\hspace{0em}%
		\frame{\includegraphics[width=0.24\textwidth]{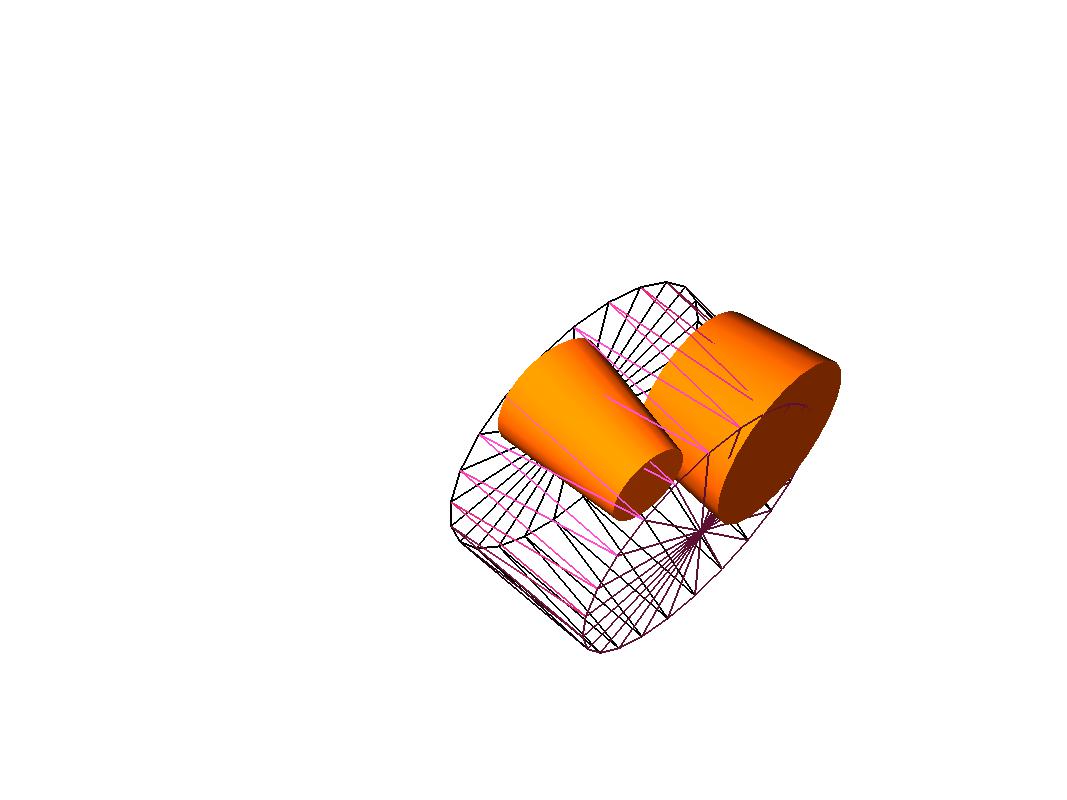}}
		\caption{Geometric Verification\label{fig:overview_verify}}
	\end{subfigure}\hspace{0em}%
	\begin{subfigure}{0.19\textwidth}
		\frame{\includegraphics[width=1\textwidth]{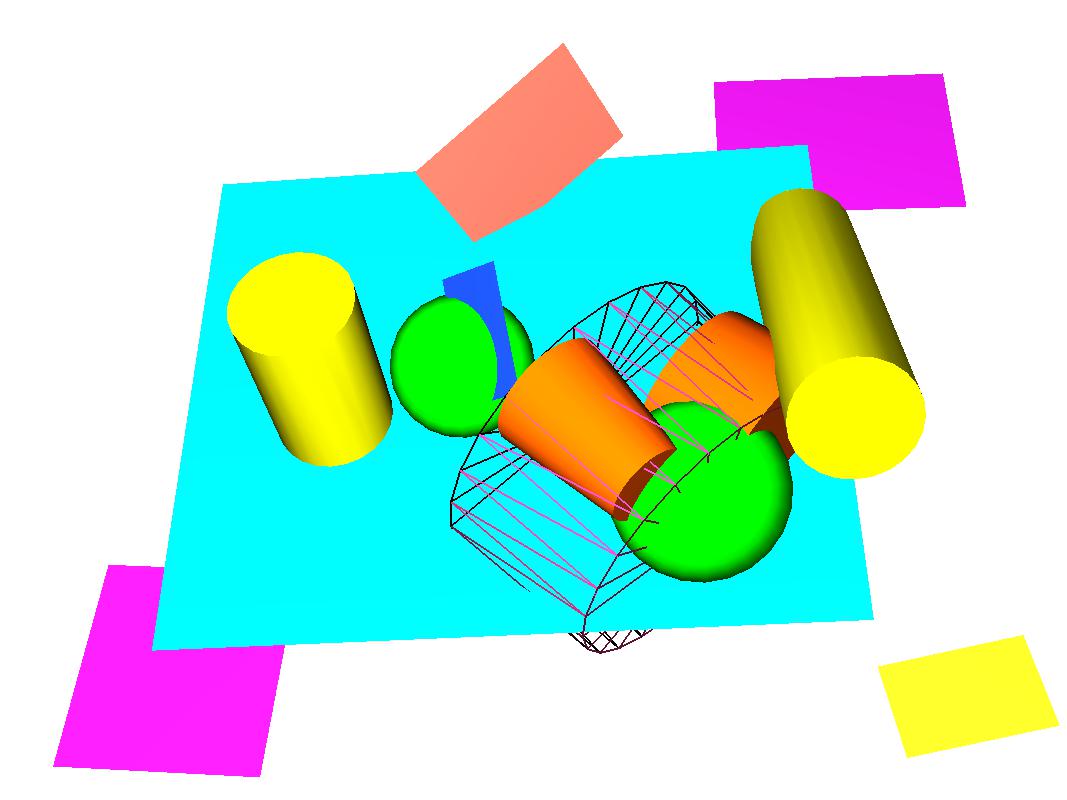}}
		\caption{Fitted Primitives\label{fig:overview_fit}}
	\end{subfigure}
	\caption{BAGSFit overview.\label{fig:overview} In~\ref{fig:overview_seg}, a proper form of a range image, e.g., its normal map, is input to a fully convolutional neural network for segmentation. We use the same visualization style for the CNN as in~\cite{yu2017casenet}, where each block means layers sharing a same spatial resolution, decreasing block height means decimating spatial resolution by a half, and red dashed lines means loss computation. The black dashed line is only applied for joint boundary detection with multi-binomial loss where low-level edge features are expected to be helpful if skip-concatenated for the final boundary classification. The resulting segmentation probability maps ${\mathbf{Y}_k}$ (top row of~\ref{fig:overview_verify}, darker for higher probability) for each primitive class $k$ are sent through a geometric verification to correct any misclassification by fitting the corresponding class of primitives (bottom row of~\ref{fig:overview_seg}). Finally, fitted primitives are shown in~\ref{fig:overview_fit}. Without loss of generality, this paper only focuses on four common primitives: plane, sphere, cylinder, and cone.}
\end{figure*}
\section{Framework Overview}

Figure~\ref{fig:overview} gives a visual overview of the multi-model primitive fitting process by our BAGSFit framework.
As introduced above, the front-end of this framework (Figure~\ref{fig:overview_seg}) mimics the human visual perception process
in that it does not explicitly use any geometric fitting error or loss in the CNN.
Instead, it takes advantage of a set of stable features learned by CNN that can robustly discriminate points belonging to different primitive classes.
The meaning of a pixel of the output probability map (top row of Figure~\ref{fig:overview_verify}) can be interpreted as
how much that point and its neighborhood look like a specific primitive class, where the neighborhood size is the CNN receptive field size.

Such a segmentation map could already be useful for more complex tasks~\cite{martens2017geometric},
yet for the sake of a robust primitive fitting pipeline, one cannot fully trust this segmentation map
as it inevitably contains misclassification, just like all other image semantic segmentations.
Fortunately, by separating pixels belonging to individual primitive classes, our original multi-model problem is converted to an easier multi-instance problem.
Following this segmentation, a geometric verification step based on efficient RANSAC~\cite{schnabel2007efficient} incorporates our strong prior knowledge, i.e., the mathematical definitions of those primitive classes, 
to find the parametric models of the objects for each type of primitives.
Note that RANSAC variants using prior inlier probability to improve sampling efficiency are not adopted in this research, because 1) they are orthogonal to the proposed pipeline; and 2) the robustness of primitive fitting is highly dependent on the spatial distribution of samples. Different from spatial consistency based methods~\cite{Sattler2009SCRAMSAC,Ni2009GroupSAC} mainly dealing with homography detection, in our 3D primitive fitting task, samples with points very close to each other usually lead to bad primitive fitting results~\cite{schnabel2007efficient}. Thus the potential of using the CNN predicted class probabilities to guide the sampling process, while being interesting, will be deferred for future investigations.

The advantage for this geometric segmentation task is that exact spatial constraints can be applied to detect correct primitives even with noisy segmentation results. One could use the inliers after geometric verification to correct the CNN segmentation results, similar to the CRF post-processing step in image semantic segmentation that usually improves segmentation performance.

\section{Ground Truth from Simulation}
Before going to the details of our segmentation CNN, we need to first address the challenge of preparing training data,
because as most state-of-the-art image semantic segmentation methods, our CNN needs to be trained by supervision.
To our best knowledge, we are the first to introduce such a geometric primitive segmentation task for CNN,
thus there is no existing publicly available datasets for this task.
For image semantic segmentation, there have been many efforts to use simulation for ground truth generation.
Yet it is hard to make CNNs trained over simulated data generalize to real world images,
due to intrinsic difficulties of tuning a large number of variables affecting the similarities between simulated images and real world ones.

However, since we are only dealing with geometric data,
and that 3D observation is less sensitive to environmental variations,
plus observation noise models of most 3D sensors are well studied,
we hypothesize that simulated 3D scans highly resemble real world ones
such that CNNs trained on simulated scans can generalize well to real world data.
If this is true, then for this geometric task, we can get infinite number of point-wise ground truth almost for free.

Although saved from tedious manual labeling, we still need a systematic way of generating
both random scene layouts of primitives and scan poses
so that simulated scans are meaningful and covers true data variation as much as possible.
Due to the popular Kinect-like scanners, which mostly applied in indoor environment,
we choose to focus on simulating indoor scenes.
And note that this dose not limit our BAGSFit framework to only indoor situations.
Given a specific type of scenes and scanners, one should be able to adjust the random scene generation protocols similarly.
Moreover, we hypothesize that the CNN is less sensitive to the overall scene layout.
What's more important is to show the CNN enough cases of different primitives occluding and intersecting with each other.

Thus, we choose to randomly generate a room-like scene with 10 meters extent at each horizontal direction.
An elevated horizontal plane representing a table top is generated at a random position near the center of the room. 
Other primitives are placed near the table top to increase the complexity. 
Furthermore, empirically, the orientation of cylinder/cone axis or plane normal is dominated by horizontal or vertical directions in real world.
Thus several primitive instances at such orientations are generated deliberately in addition to fully random ones.
For planes, two additional disk shaped planes are added to make the dataset more general.
To make the training set more realistic, two NURBS surfaces (class name ``Other'' in Figure~\ref{fig:teaser}) are added,
representing objects not explained by our primitive library in reality.

\begin{figure} 
	\centering
	\includegraphics[width=.5\columnwidth]{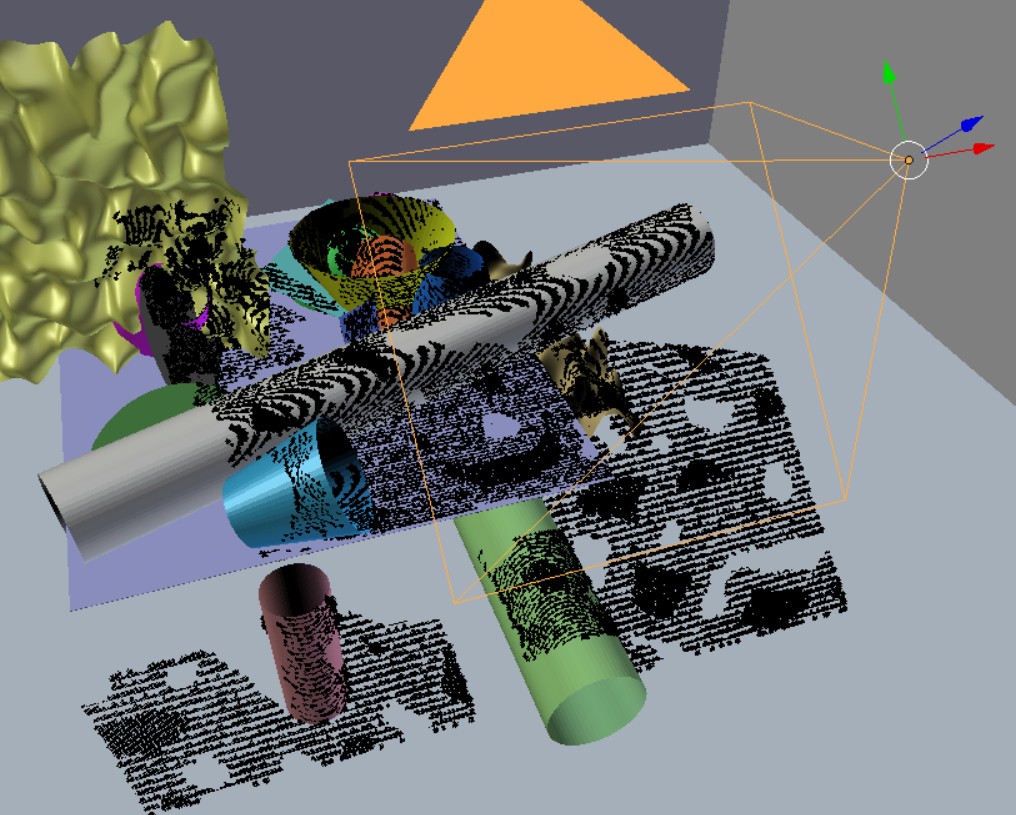}
	\caption{A simulated Kinect scan of a random scene. Black dots represents the scanned points. \label{fig::blensor_scan}}
\end{figure}

An existing scanner simulator, Blensor~\cite{gschwandtner2011blensor}, was used to simulate VGA-sized Kinect-like scans,
where class and instance IDs can be easily obtained during the virtual scanning process by ray-tracing.
The default Kinect scanner was adopted except that the noise sigma parameter was set to 0.005.
Note that we do not carefully tune the parameters to match the simulated noise with real Kinect noise model.
In fact, our simulated scanner produces slightly more noisy points than and a real Kinect sensor.
To generate random scan poses, the virtual scanners were firstly placed around the center of the ``table''. 
Then camera viewing directions were sampled on a grid of longitudinal $\pi/6$
and latitudinal $\pi/12$ intervals ranging from $[-\pi, \pi)$ and $[-\pi/6, \pi/2)$, resulting in $81$ directions in total.
For each direction, two distances to the table's center ranging between $[1.5,4]$m were uniformly sampled. 
Thus, for each scene we obtain a total number of 192 scan poses.
At last, a uniform noise between $[-\pi/24, \pi/24]$ was added to each viewing direction both horizontally and vertically. 
Figure~\ref{fig::blensor_scan} shows the screenshot of such a scan.
Totally 20 scenes were generated following this protocol.
18 scenes, i.e. 3456 scans, were split for training,
and the other 2 scenes, i.e. 384 scans, were used for validation.
The test set is generated through a similar protocol, containing 20 scenes (each with 36 scans).
Note that invalid points were converted to the zero-depth point avoiding computation issues.

\section{Boundary Aware Geometric Segmentation}\label{sec:bias}
Our segmentation network (Figure~\ref{fig:overview_seg}) follows the same basic network as described in~\cite{yu2017casenet},
which is based on the 101-layer ResNet~\cite{He2016} with minor modifications to improve segmentation performance.
While the semantic segmentation CNN architecture is actively being developed,
there are several design choices to be considered to achieve the best performance on a given base network for our new task.

\textbf{Position vs. Normal Input}.
The first design choice is about the input representation.
Since we are dealing with 3D geometric data, what form of input should be supplied to the CNN?
A naive choice is to directly use point positions as a 3-channel tensor input.
After all, this is the raw data we get in reality, and if the CNN is powerful enough,
it should be able to learn everything from this input form.
However, it is unclear how or whether necessary to normalized it.

A second choice is to use estimated per-point unit normals as the input.
This is also reasonable, because we can almost perceive the correct segmentation by just looking as the normal maps as shown in Figure~\ref{fig:overview_seg}.
Plus it is already normalized, which usually enables better CNN training.
However, since normals are estimated from noisy neighboring points, one might have concerns about loss of information compared with the previous choice.
And a third choice is to combine the first two, resulting in a 6-channel input, through which one might hope the CNN to benefit from merits of both.

\textbf{Multinomial vs. Multi-binomial Loss}.
The second design question is: what kind of loss function to use?
While many semantic segmentation CNN choose the multinomial cross-entropy loss through a softmax function,
recent studies have found other loss functions such as the self-balancing multi-binomial loss~\cite{yu2017casenet} to perform better for certain tasks,
with weights accounting for imbalanced classes.
In this study, we consider two types of loss functions: 1) the classic ``softmax loss'',
and 2) a multi-binomial loss with class-specific loss weights $\beta_k$ as hyper-parameters:
\begin{align}\label{eq:loss}
\mathcal{L}(\mathbf{W})= & \sum_{k} \beta_k \mathcal{L}_k(\mathbf{W}) \\ \nonumber
= & \sum_{k} \beta_k \sum_{\mathbf{p}} \{ - \mathbf{\bar{Y}}_k(\mathbf{p})\log \mathbf{Y}_k(\mathbf{p}|\mathbf{I};\mathbf{W}) \\ \nonumber
& - (1-\mathbf{\bar{Y}}_k(\mathbf{p}))\log(1-\mathbf{Y}_k(\mathbf{p}|\mathbf{I};\mathbf{W}))\},
\end{align}
where $\mathbf{W}$ are the learnable parameters, $\mathbf{p}$ a pixel index, $\mathbf{\bar{Y}}_k$ the ground truth binary image and $\mathbf{Y}_k$ the network predicted probability map of the $k$-th primitive class ($k\in[1,K]$), and $\mathbf{I}$ the input data.
We set $\beta_k$ to be proportional to 1 over the total number of $k$-th class points in the training set.

\textbf{Separate vs. Joint Boundary Detection}.
When multiple instances of a same primitive class occlude or intersect with each other,
even an ideal primitive class segmentation can not divide them into individual segments,
leaving a multi-instance fitting problem still undesirable for the geometric verification step to solve,
which discounts the original purpose of this geometric segmentation.
Moreover, boundaries usually contains higher noises in terms of estimated normals,
which could negatively affect primitive fittings that use normals (e.g., 2-point based cylinder fitting).
One way to alleviate the issue is to cut such clusters into primitive instances by instance-aware boundaries.
To realize this, we also have two choices, 1) training a separate network only for instance boundary detection,
or 2) treating boundary as an additional class to be segmented jointly with primitive classes.
One can expect the former to have better boundary detection results as the network focuses to learn boundary
features only, although as a less elegant solution with more parameters and longer running time.
Thus it is reasonable to trade the performance a bit for the latter one.
Note that with such a step, we could already move from category- to boundary- and thus instance-aware segmentation by region-grow after removing all instance-aware boundaries.

\textbf{Handling of Background Class}.
When generating random scenes, we added NURBS modeling background points not explained by the four primitive classes,
for a more realistic and challenging dataset.
Thus we need to properly handle them in the CNN.
Should we ignore background class when computing the loss, or add it as an additional class?

For all of the above design questions, we will rely on experiments to empirically select the best performing ones.

\section{Geometric Verification and Evaluation}

\subsection{Verification by Fitting} \label{sec:verify_by_fit}
Given the predicted probability maps $\{\mathbf{Y}_k\}$, we need to generate and verify primitive hypotheses and
fit primitive parameters of the correct ones to complete our mission.

One direct way of hypothesis generation is to simply binarize the BAGS output $\{\mathbf{Y}_k\}$ by thresholding
to produce a set of connected components, and fit only one $k$-th class primitive for a component coming from ${\mathbf{Y}_k}$.
However, when the CNN incorrectly classify certain critical regions due to non-optimal thresholds, two instances can be connected,
thus leading to suboptimal fittings or miss detection of some instances.
Moreover, a perfect BAGS output may bring another issue that an instance gets cut into several smaller pieces due to occlusions
(e.g., the top left cylinder in Figure~\ref{fig:overview_seg}).
And fitting in smaller regions of noisy scans usually result in false instance rejection or lower estimation accuracy.
since the core contribution of this paper is to propose and study the feasibility of BAGSFit as a new strategy towards this problem,
we leave it as our future work to develop more systematic ways to better utilize $\{\mathbf{Y}_k\}$ for primitive fitting.

In this work, we simply follow a classic ``$\arg\max$'' prediction on $\{\mathbf{Y}_k\}$ over each point,
and get $K$ groups of hypothesis points associated to each of the $K$ primitive classes.
Then we solve $K$ times of multi-instance primitive fitting using the RANSAC-based method~\cite{schnabel2007efficient}.
This is more formally described in Algorithm~\ref{alg::prim_fit}.
Note this does not completely defeat the purpose of BAGS.
The original RANSAC-based method feed the whole point cloud into the pipeline and detect primitives sequentially in a greedy manner.
Because it tends to detect larger objects first, smaller primitives close to large ones could often be missed,
as their member points might be incorrectly counted as inlier of larger objects,
especially if the inlier threshold is improperly set.
BAGS can alleviate such effects and especially removing boundary points from RANSAC sampling is expected to improve its performance.

\begin{algorithm}
	\caption{Primitive Fitting from $\arg\max$ Hypotheses}\label{euclid}\label{alg::prim_fit}
	\begin{algorithmic}
		\Function{PrimitiveFitting}{$\mathbf{I}, \{\mathbf{Y}_k\}$}
			\State $\mathbf{M}_k \gets \emptyset, \forall k\in[1,K]$  	\Comment{initialize hypotheses sets}
			\For {$\mathbf{p}\in\mathbf{I}$}							\Comment{assign a pixel to its best set}
				\State $j=\arg\max_k\{\mathbf{Y}_k(\mathbf{p})\}$
				\State $\mathbf{M}_{j} \gets \mathbf{M}_{j} \bigcup \{\mathbf{p}\}$
			\EndFor
			\State $Prims \gets \emptyset$								
			\For {$k \in [1, K]$}										\Comment{detect primitives from each set}
				\State $Prims \gets Prims \bigcup \text{EfficientRANSAC}(\mathbf{M}_k, \mathbf{I})$
			\EndFor
		\State \textbf{return} $Prims$
		\EndFunction
	\end{algorithmic}
\end{algorithm}

\subsection{Primitive Fitting Evaluation}
It is non-trivial to design a proper set of evaluation criteria for primitive detection and fitting accuracy,
and we are not aware of any existing work or dataset that does so.
It is difficult to comprehensively evaluate and thus compare different primitive fitting methods
partly because 1) as mentioned previously, due to occlusion, a single instance are commonly fitted into multiple primitives,
both of which may be close enough to the ground truth instance;
and 2) such over detection might also be caused by improper inlier thresholds on a noisy data. 

Pixel-wise average precision (AP) and AP of instances matched at various levels (50$\sim$90\%) of point-wise intersection-over-union (IoU)
are used for evaluating image based instance segmentation problems~\cite{bai2016deep}. 
However, this typical IoU range is inappropriate for our problem.
More than 50\% IoU means at most one fitted primitive can be matched for each true instance.
Since we don't need more than 50\% of true points to fit a reasonable primitive representing the true one,
this range is over-strict and might falsely reject many good fits:
either more than 50\% true points are taken by other incorrect fits,
or during observation the true instance is occluded and split into pieces each containing less than 50\% true points (see Figure~\ref{fig:simu} for more examples).
After all, a large IoU is not necessary for good primitive fitting.

Thus, the IoU is replaced by intersection-over-true (IoT) in this problem. 
It indicates the number of true inliers of a predicted primitive over the total number of points in the true instance.
Thus, a predicted primitive and a true instance is matched iff 1) IoT$>$30\%
and 2) the predicted primitive having the same class as the true instance.
This indicates that one instance can have at most 3 matched predictions.

Based on the above matching criteria, a matched instance (if exists) can be identified for each predicted primitive.
On the contrary, each true instance may have several best matching prediction candidates. 
To eliminate the ambiguity, the candidate that has the smallest fit error is selected as the best match.
To be fair and consistent, fitting error is defined as the mean distance to a primitive by projecting all of the points in the true instance onto the predicted primitive. 
After the matches are found, \textit{primitive average precision} (PAP) and \textit{primitive average recall} (PAR) are used to quantify the primitive detection quality.
\begin{equation}\label{eq:pap_par}
PAP = N_{p2t}/N_p, PAR = N_{t2p}/N_t,
\end{equation}
where $N_{p2t}$ is the number of predictions having a matched true instance, $N_p$ the total number of predicted primitives, $N_{t2p}$ the number of true instance with a best prediction, and $N_t$ the total number of true instances, all counted over the whole test set.

\section{Experiments and Discussion}

\subsection{Geometric Segmentation Experiments}

\begin{table*}[t!]
	\centering
	\caption{\label{tab:seg-eval-t} Geometric segmentation evaluation. Red highlights the best along a column, while magenta for the top 3 best.}
	\resizebox{\textwidth}{!}{%
		\begin{tabular}{|l|c|ccccc|c|ccccc|c|ccccc|c|ccccc|c|} 
			\hline
			\multirow{2}{*}{}&\multicolumn{6}{c|}{Precision}&\multicolumn{6}{c|}{Recall}&\multicolumn{6}{c|}{IoU}&\multicolumn{6}{c|}{F1}&\multirow{2}{*}{Accuracy}\\
			\cline{2-25}&BND&PLN&SPH&CYL&CON&AVE&BND&PLN&SPH&CYL&CON&AVE&BND&PLN&SPH&CYL&CON&AVE&BND&PLN&SPH&CYL&CON&AVE&\\
			\hline
			N+BO&0.944&&&&&&0.820&&&&&&0.781&&&&&&0.877&&&&&&0.964\\
			\hline
			P&&0.915&0.811&0.867&0.642&0.809&&0.971&0.620&0.715&0.664&0.743&&0.891&0.599&0.655&0.488&0.658&&0.939&0.664&0.762&0.611&0.744&0.871\\
			N&&0.979&0.915&0.934&0.727&\textbf{\textcolor{magenta}{0.889}}&&\textbf{\textcolor{red}{0.988}}&\textbf{\textcolor{red}{0.884}}&0.788&\textbf{\textcolor{red}{0.829}}&\textbf{\textcolor{magenta}{0.872}}&&\textbf{\textcolor{red}{0.968}}&\textbf{\textcolor{red}{0.860}}&0.752&0.633&\textbf{\textcolor{magenta}{0.803}}&&\textbf{\textcolor{red}{0.983}}&\textbf{\textcolor{red}{0.894}}&0.826&0.734&0.859&0.924\\
			PN&&0.978&0.913&0.919&0.710&0.880&&0.984&0.868&0.806&0.797&0.864&&0.962&0.847&0.758&0.601&0.792&&0.980&0.882&0.838&0.711&0.853&0.920\\
			\hline
			P+MB&&0.929&0.818&0.888&0.658&0.823&&0.967&0.656&0.730&0.706&0.765&&0.900&0.626&0.677&0.518&0.680&&0.945&0.690&0.774&0.638&0.762&0.881\\
			N+MB&&0.978&0.899&0.923&0.737&0.884&&0.985&0.864&0.806&0.816&0.868&&0.964&0.835&0.764&0.638&\textbf{\textcolor{magenta}{0.800}}&&0.981&0.873&0.836&0.738&0.857&\textbf{\textcolor{magenta}{0.927}}\\
			PN+MB&&0.979&0.911&0.900&0.677&0.867&&0.949&0.860&0.792&0.805&0.852&&0.930&0.839&0.737&0.576&0.771&&0.958&0.875&0.817&0.686&0.834&0.894\\
			\hline
			N+BAGS&0.868&0.963&0.908&0.926&\textbf{\textcolor{red}{0.756}}&\textbf{\textcolor{magenta}{0.888}}&0.849&0.976&0.874&\textbf{\textcolor{red}{0.833}}&0.821&\textbf{\textcolor{magenta}{0.871}}&0.752&0.941&0.848&\textbf{\textcolor{red}{0.790}}&\textbf{\textcolor{red}{0.654}}&0.797&0.858&0.969&0.884&\textbf{\textcolor{red}{0.859}}&\textbf{\textcolor{red}{0.755}}&\textbf{\textcolor{magenta}{0.865}}&0.918\\
			N+MB+BAGS&0.868&0.950&0.886&0.891&0.677&0.851&0.809&0.977&0.855&0.765&0.749&0.831&0.720&0.929&0.820&0.703&0.549&0.744&0.837&0.962&0.862&0.792&0.662&0.823&0.887\\
			\hline
			N5&&\textbf{\textcolor{red}{0.980}}&\textbf{\textcolor{red}{0.917}}&\textbf{\textcolor{red}{0.940}}&0.744&\textbf{\textcolor{magenta}{0.895}}&&0.979&0.877&0.809&0.808&\textbf{\textcolor{magenta}{0.868}}&&0.960&0.854&0.776&0.642&\textbf{\textcolor{magenta}{0.808}}&&0.979&0.889&0.844&0.741&\textbf{\textcolor{magenta}{0.863}}&\textbf{\textcolor{magenta}{0.940}}\\
			N5+MB&&0.978&0.911&0.920&0.725&0.884&&0.977&0.862&0.804&0.793&0.859&&0.956&0.841&0.760&0.614&0.793&&0.977&0.878&0.834&0.719&0.852&\textbf{\textcolor{magenta}{0.932}}\\
			N5+BAGS&0.847&0.966&0.906&0.932&0.728&0.883&0.804&0.970&0.873&0.808&0.812&0.853&0.702&0.939&0.845&0.769&0.630&0.777&0.825&0.968&0.883&0.842&0.732&0.850&0.921\\
			\hline
		\end{tabular}
	}
	\centering
\end{table*}

\begin{figure*}[t!] 
	\centering

	\frame{\includegraphics[width=.124\textwidth]{{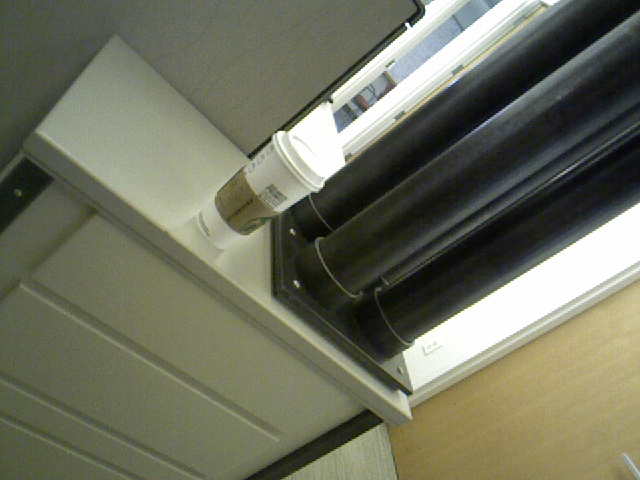}}}\hspace{0em}%
	\frame{\includegraphics[width=.124\textwidth]{{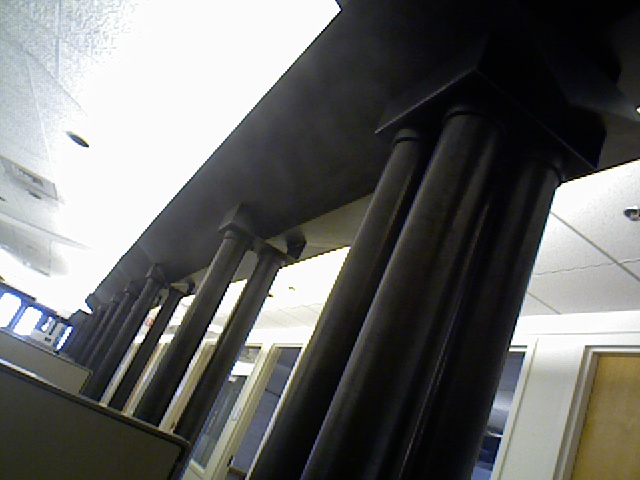}}}\hspace{0em}%
	\frame{\includegraphics[width=.124\textwidth]{{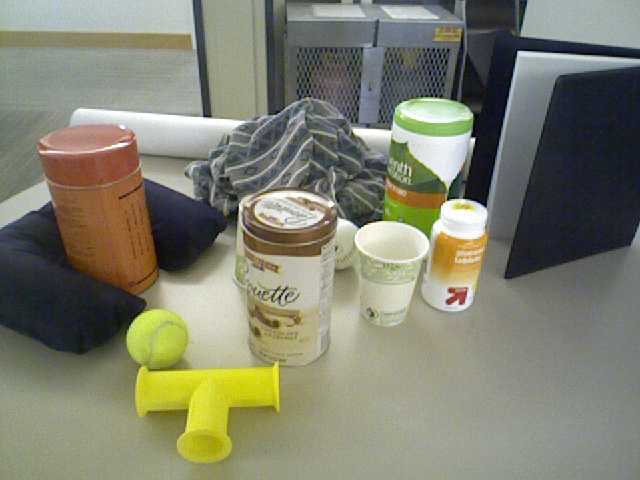}}}\hspace{0em}%
	\frame{\includegraphics[width=.124\textwidth]{{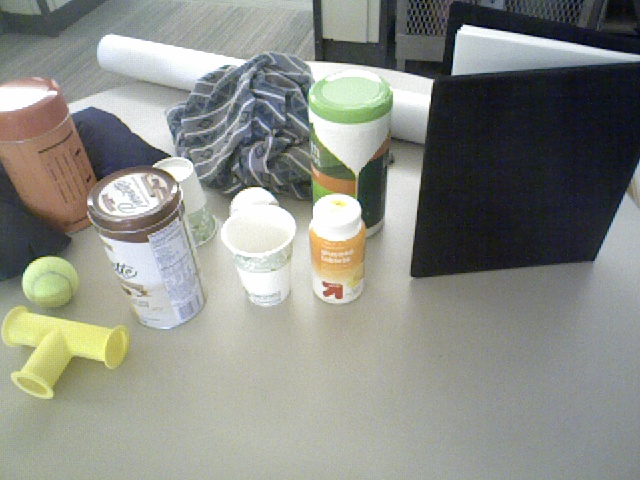}}}\hspace{0em}%
	\frame{\includegraphics[width=.124\textwidth]{{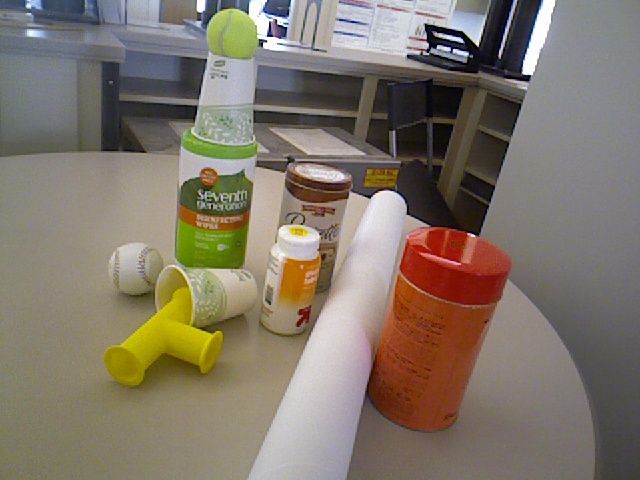}}}\hspace{0em}%
	\frame{\includegraphics[width=.124\textwidth]{{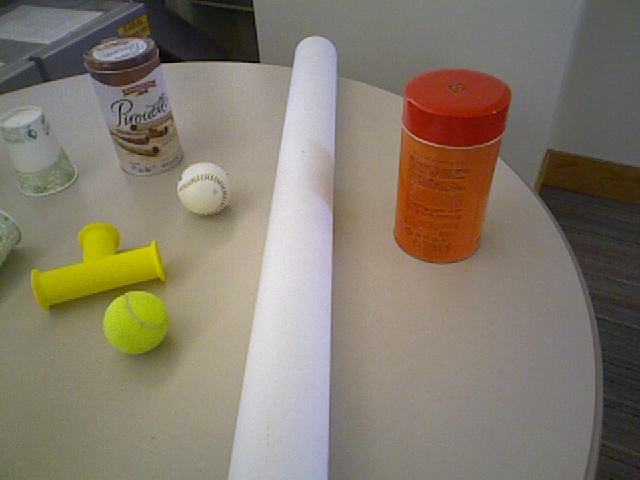}}}\hspace{0em}%
	\frame{\includegraphics[width=.124\textwidth]{{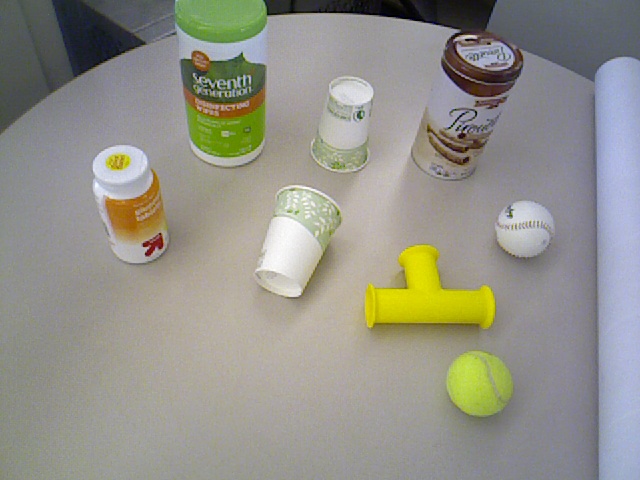}}}\hspace{0em}%
	\frame{\includegraphics[width=.124\textwidth]{{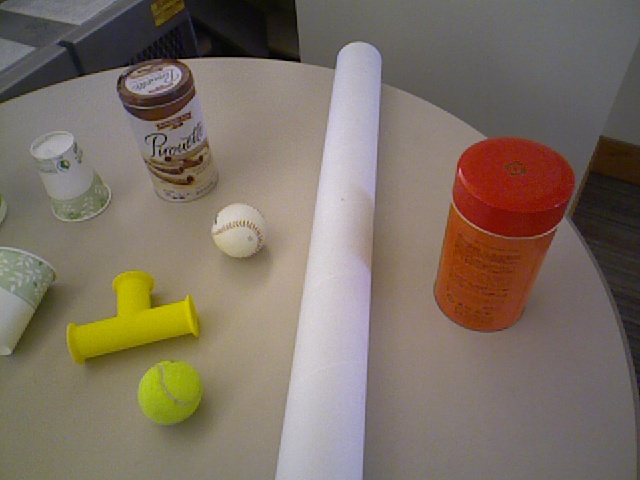}}}
	
	\frame{\includegraphics[width=.124\textwidth]{{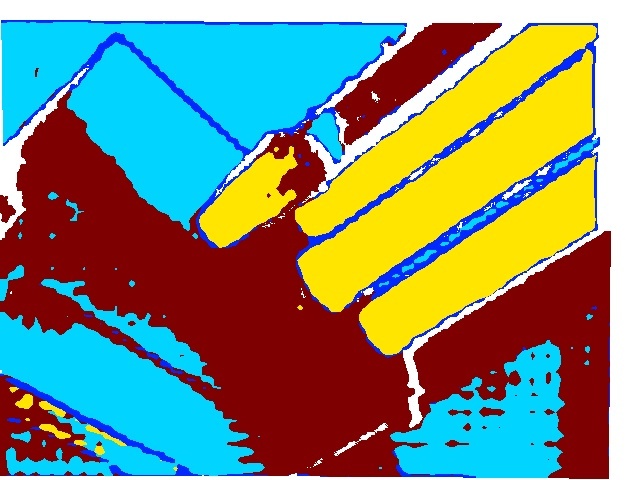}}}\hspace{0em}%
	\frame{\includegraphics[width=.124\textwidth]{{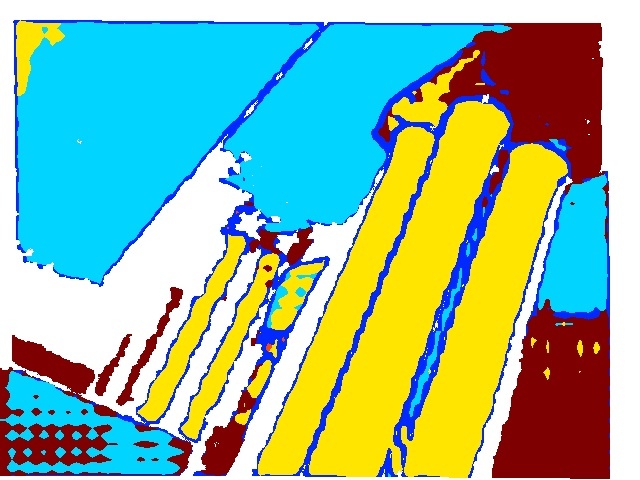}}}\hspace{0em}%
	\frame{\includegraphics[width=.124\textwidth]{{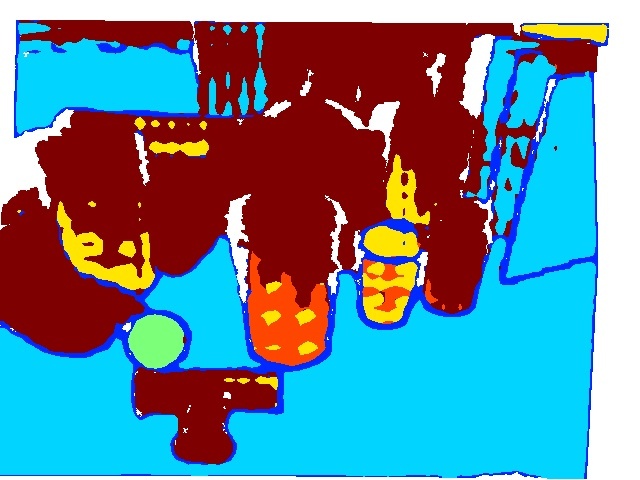}}}\hspace{0em}%
	\frame{\includegraphics[width=.124\textwidth]{{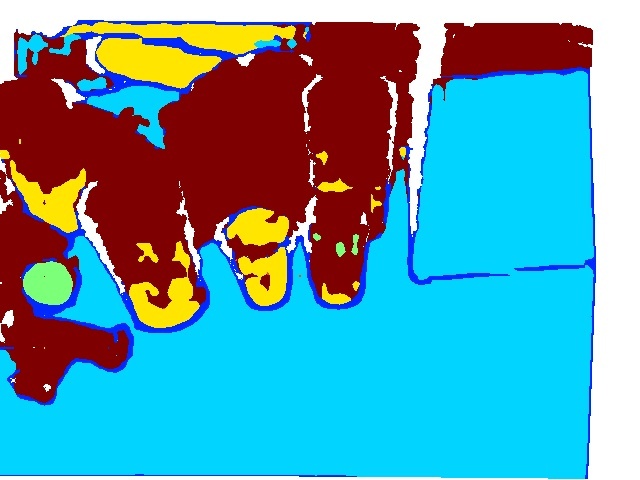}}}\hspace{0em}%
	\frame{\includegraphics[width=.124\textwidth]{{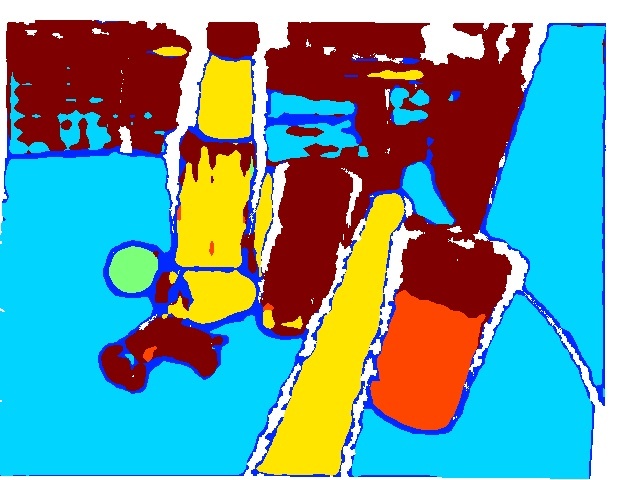}}}\hspace{0em}%
	\frame{\includegraphics[width=.124\textwidth]{{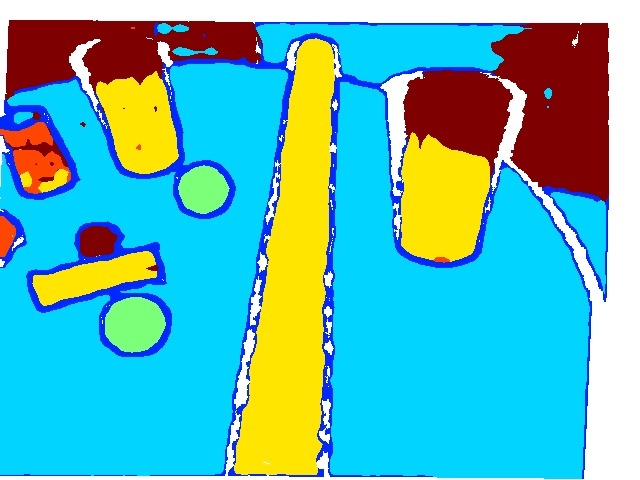}}}\hspace{0em}%
	\frame{\includegraphics[width=.124\textwidth]{{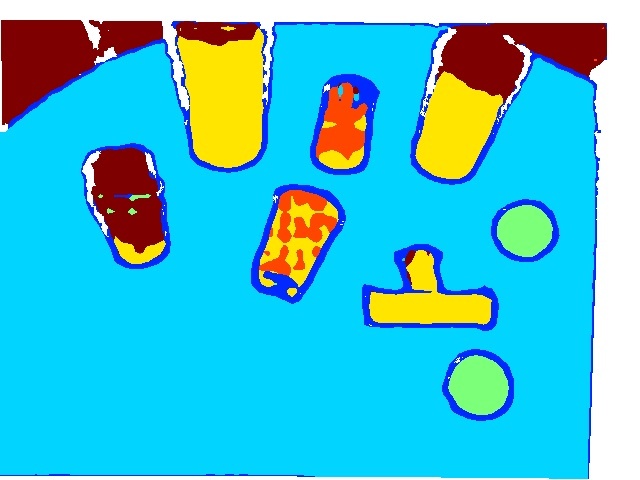}}}\hspace{0em}%
	\frame{\includegraphics[width=.124\textwidth]{{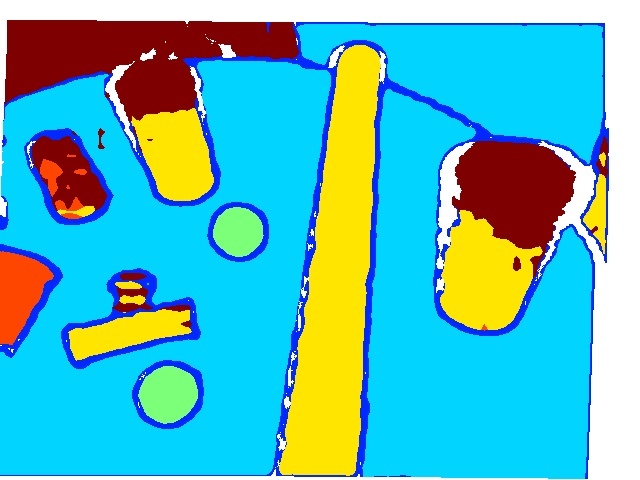}}}
	
	\frame{\includegraphics[width=.124\textwidth]{{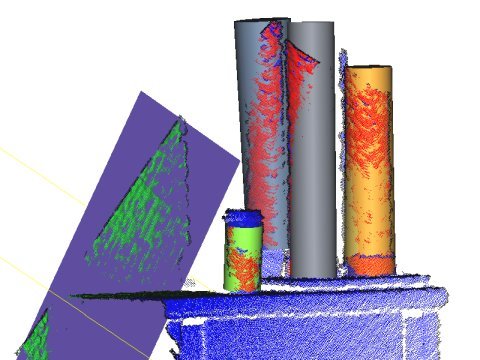}}}\hspace{0em}%
	\frame{\includegraphics[width=.124\textwidth]{{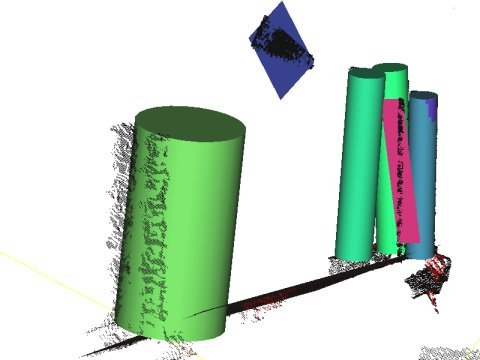}}}\hspace{0em}%
	\frame{\includegraphics[width=.124\textwidth]{{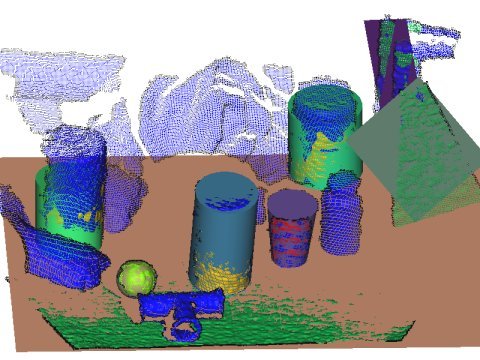}}}\hspace{0em}%
	\frame{\includegraphics[width=.124\textwidth]{{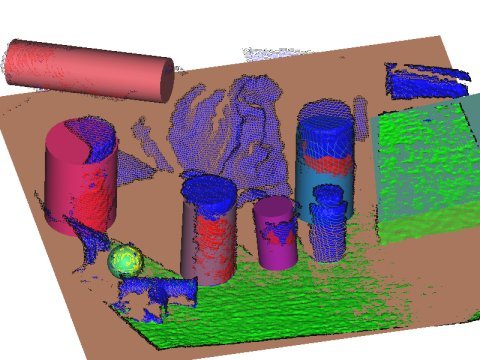}}}\hspace{0em}%
	\frame{\includegraphics[width=.124\textwidth]{{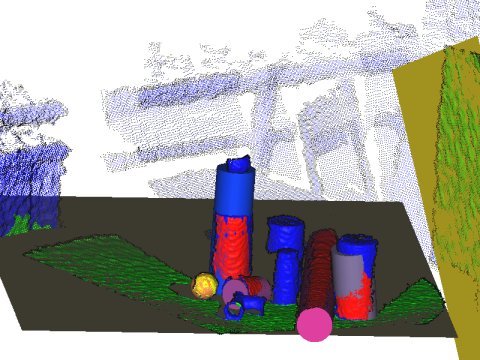}}}\hspace{0em}%
	\frame{\includegraphics[width=.124\textwidth]{{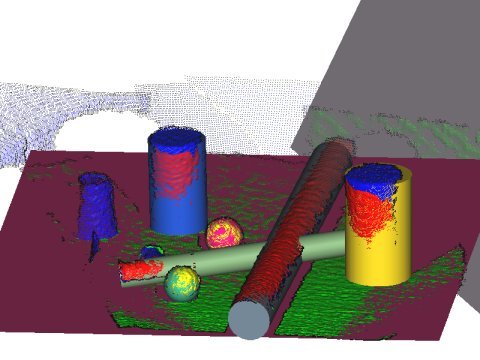}}}\hspace{0em}%
	\frame{\includegraphics[width=.124\textwidth]{{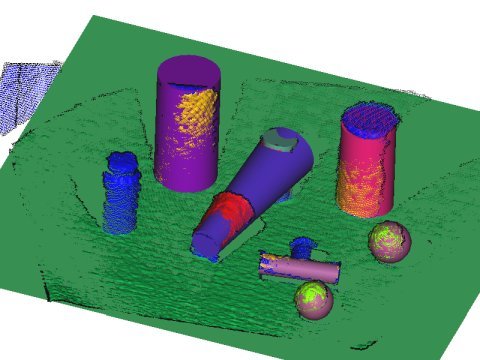}}}\hspace{0em}%
	\frame{\includegraphics[width=.124\textwidth]{{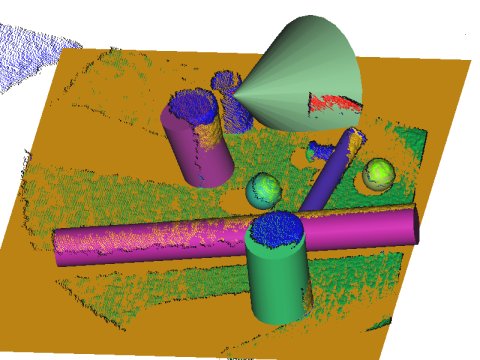}}}
	
	\caption{BAGSFit (\textbf{N5+BAGS}) on real Kinect scans. Top: RGB image of the scanned scene. Middle: segmentation results. Bottom: fitted primitives (randomly colored) rendered together with real scans.\label{fig:real}}
\end{figure*}

\textbf{Network Short Names}. To explore answers to the design questions raised in section~\ref{sec:bias}, we designed several CNNs and their details with short names are listed as follows:
\begin{itemize}
	\item \textbf{P/N/PN}. Basic networks, using position (\textbf{P}), normal (\textbf{N}), or both (\textbf{PN}) as input, trained with a multinomial loss function, outputting a 4-channel \textit{mutual-exclusive} class probability maps (i.e., each pixel's probabilities sum up to one, $K=4$). Background class points, the NURBS, are ignored for loss computation.
	\item \textbf{P/N/PN+MB}. Same as the above basic networks except trained using the multi-binomial (\textbf{MB}) loss function as in equation~\eqref{eq:loss}, outputting a 4-channel \textit{non-mutual-exclusive} class probability maps (i.e., each pixel's probabilities not necessarily sum up to one, thus being multi-binomial classifiers, $K=4$).
	\item \textbf{N+BAGS}. Network trained with normal input and BAGS labels (i.e., instance-aware boundary as an additional class jointly trained, $K=5$).
	\item \textbf{N+MB+BAGS}. Same as \textbf{N+BAGS} except trained using a multi-binomial manner ($K=5$).
	\item \textbf{N5}. Same as basic network \textbf{N} except treating the background class as an additional class involved in loss computation ($K=5$).
	\item \textbf{N5+MB}. Same as \textbf{N5} except trained using a multi-binomial manner ($K=5$).
	\item \textbf{N5+BAGS}. Same as \textbf{N+BAGS} except trained using a multi-binomial manner (i.e., boundary and NURBS are two additional classes jointly trained, $K=6$).
	\item \textbf{N+BO}. Same as \textbf{N} except only trained to detect boundary (i.e., a binary classifier, $K=2$).
\end{itemize} 

\textbf{Implementation Details}.
We implemented the geometric segmentation CNNs using \textit{Caffe}~\cite{jia2014caffe} and \textit{DeepLabv2}~\cite{chen2016deeplab}.
Normals were estimated by PCA using a $5\times 5$ window.
We use meters as the unit for networks requiring position input.
Instance-aware boundaries were calculated if not all pixels belong to a same instance (or contain invalid points) in a $5\times 5$ window.
Input data size was randomly cropped into $440\times 440$ during training time, while full VGA resolution was used during test time.
All of our networks were trained with the following hyper-parameters tuned on the validation set:
50 training epochs (i.e. 17280 iterations), batch size 10,
learning rate 0.1 linearly decreasing to zero until the end of training,
momentum 0.9, weight decay 5e-4.
The networks were trained and evaluated using several NVIDIA TITAN X GPUs each with 12 GB memory, with a 2.5 FPS testing frame rate.

\begin{table*}[t!]
	\centering
	\caption{\label{tab:seg-fit-t}Primitive fitting evaluation.  Red highlights the best along a column, while magenta highlights the top 3 best.}
	\resizebox{\textwidth}{!}{%
		\begin{tabular}{|l|ccccc|ccccc|c|ccccc|ccccc|ccccc|} 
			\hline
			\multirow{2}{*}{}&\multicolumn{5}{c|}{No. Primitives Fitted ($N_p$)}&\multicolumn{5}{c|}{No. Matched Instance($N_{t2p}$) }&\multirow{2}{*}{$\frac{N_{t2p}^{all}}{N_p^{all}}$}&\multicolumn{5}{c|}{Primitive Average Precision (PAP)}&\multicolumn{5}{c|}{Primitive Average Recall (PAR)}&\multicolumn{5}{c|}{Fitting Error (cm)}\\
			\cline{2-11}\cline{13-27}&PLN&SPH&CYL&CON&ALL&PLN&SPH&CYL&CON&ALL&&PLN&SPH&CYL&CON&ALL&PLN&SPH&CYL&CON&ALL&PLN&SPH&CYL&CON&ALL\\
			\hline
			ERANSAC&4596&1001&2358&3123&11078&2017&542&942&879&4380&0.395&0.453&0.541&0.402&0.286&0.403&0.500&0.432&0.403&0.443&0.456&0.915&0.324&\textcolor{red}{0.766}&0.954&0.810\\
			\hline
			P&5360&621&2242&2037&10260&2448&591&1219&944&5202&0.507&0.470&0.952&0.549&0.468&0.516&0.607&0.471&0.521&0.476&0.541&0.936&\textcolor{red}{0.248}&0.931&0.519&\textbf{\textcolor{magenta}{0.759}}\\
			N&4617&961&2789&2492&10859&\textbf{\textcolor{red}{2565}}&\textbf{\textcolor{red}{870}}&1456&1254&\textbf{\textcolor{magenta}{6145}}&0.566&0.571&0.905&0.532&0.507&0.576&\textbf{\textcolor{red}{0.636}}&\textbf{\textcolor{red}{0.693}}&0.623&\textbf{\textcolor{red}{0.633}}&\textbf{\textcolor{magenta}{0.640}}&0.903&0.403&1.229&0.657&0.866\\
			PN&4537&888&3172&2133&10730&2522&859&\textbf{\textcolor{red}{1498}}&1197&\textbf{\textcolor{magenta}{6076}}&0.566&0.572&0.967&0.480&0.570&0.577&0.625&0.684&\textbf{\textcolor{red}{0.641}}&0.604&\textbf{\textcolor{magenta}{0.632}}&0.903&0.397&1.196&0.628&0.852\\
			\hline
			P+MB&5103&654&2201&2061&10019&2373&625&1249&1010&5257&0.525&0.479&0.956&0.573&0.493&0.534&0.588&0.498&0.534&0.510&0.547&0.892&0.283&0.996&0.549&0.767\\
			N+MB&4479&931&2732&2654&10796&2528&857&1442&1222&\textbf{\textcolor{magenta}{6049}}&0.560&0.581&0.921&0.536&0.466&0.571&0.627&0.682&0.617&0.617&\textbf{\textcolor{magenta}{0.630}}&0.896&0.397&1.187&0.719&0.865\\
			PN+MB&4236&951&3169&2305&10661&2427&856&1455&1168&5906&0.554&0.589&0.900&0.467&0.516&0.565&0.602&0.682&0.622&0.589&0.615&0.873&0.394&1.193&0.699&0.852\\
			\hline
			N+BAGS&3893&845&2299&2108&9145&2279&796&1453&1149&5677&0.621&0.594&0.942&0.637&0.548&0.626&0.565&0.634&0.621&0.580&0.591&0.765&0.363&1.144&0.587&0.768\\
			N+MB+BAGS&3815&800&2599&1947&9161&2249&775&1356&1002&5382&0.587&0.598&0.969&0.528&0.518&0.594&0.558&0.617&0.580&0.506&0.560&0.754&0.357&1.149&\textbf{\textcolor{red}{0.533}}&\textbf{\textcolor{magenta}{0.753}}\\
			\hline
			N5&3701&863&1874&1876&8314&2490&859&1458&\textbf{\textcolor{red}{1226}}&6033&\textbf{\textcolor{magenta}{0.726}}&\textbf{\textcolor{red}{0.693}}&0.995&0.793&\textbf{\textcolor{red}{0.663}}&\textbf{\textcolor{magenta}{0.740}}&0.617&0.684&0.624&0.619&0.628&0.841&0.395&1.163&0.617&0.815\\
			N5+MB&3717&858&1920&1930&8425&2490&857&1479&1199&6025&\textbf{\textcolor{magenta}{0.715}}&0.689&0.999&0.783&0.634&\textbf{\textcolor{magenta}{0.729}}&0.617&0.682&0.633&0.605&0.627&0.842&0.398&1.148&0.659&0.821\\
			N5+BAGS&3500&804&1765&1730&7799&2254&804&1397&1129&5584&\textbf{\textcolor{magenta}{0.716}}&0.654&\textbf{\textcolor{red}{1.000}}&\textbf{\textcolor{red}{0.796}}&0.658&\textbf{\textcolor{magenta}{0.723}}&0.559&0.640&0.598&0.570&0.581&\textbf{\textcolor{red}{0.742}}&0.367&1.096&0.555&\textbf{\textcolor{magenta}{0.740}}\\
			\hline
			MaskRCNN&2422&890&1771&3728&8811&1104&553&944&1017&3618&0.411&0.526&0.621&0.590&0.282&0.445&0.274&0.440&0.404&0.513&0.377&10.789&35.361&19.122&23.348&18.619\\\hline
		\end{tabular}
	}
	\centering
\end{table*}

\begin{figure*}[t!] 
	\centering
	\frame{\includegraphics[width=.11\textwidth]{{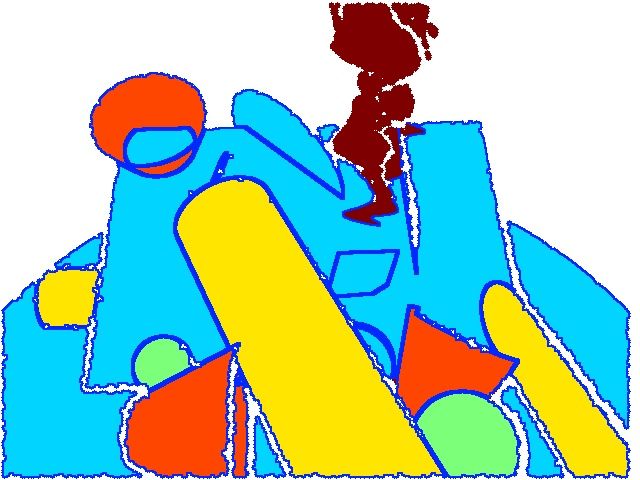}}}\hspace{0em}%
	\frame{\includegraphics[width=.11\textwidth]{{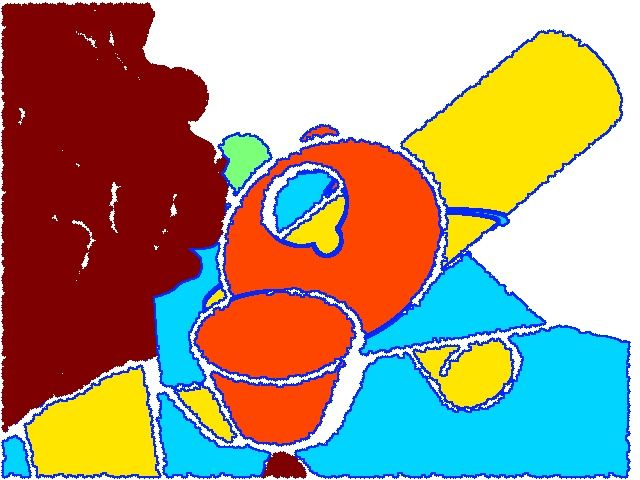}}}\hspace{0em}%
	\frame{\includegraphics[width=.11\textwidth]{{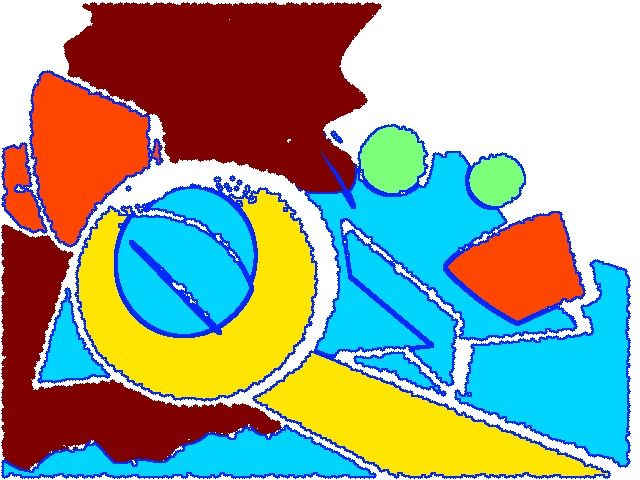}}}\hspace{0em}%
	\frame{\includegraphics[width=.11\textwidth]{{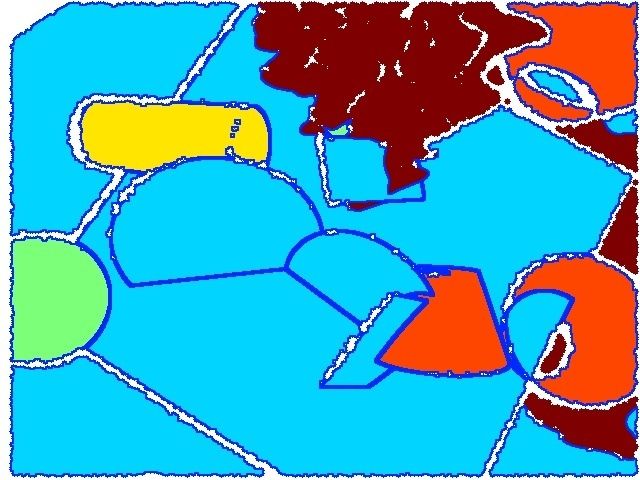}}}\hspace{0em}%
	\frame{\includegraphics[width=.11\textwidth]{{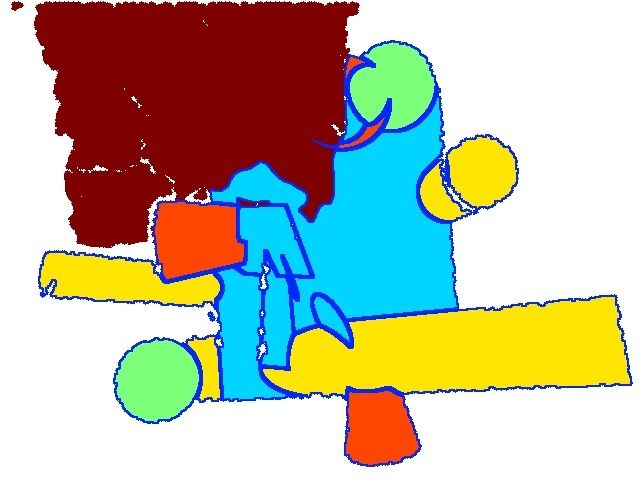}}}\hspace{0em}%
	\frame{\includegraphics[width=.11\textwidth]{{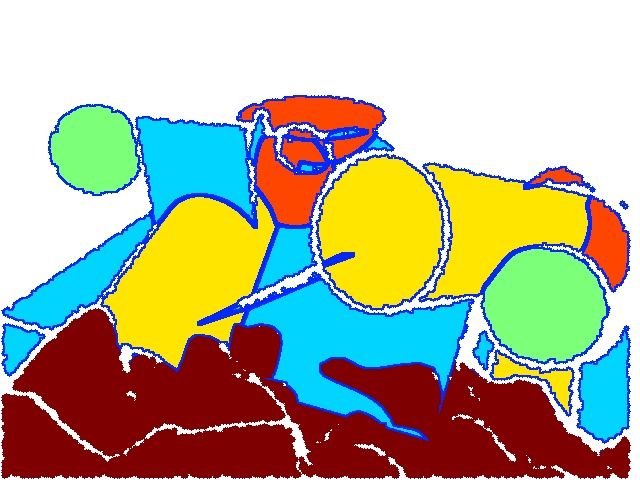}}}\hspace{0em}%
	\frame{\includegraphics[width=.11\textwidth]{{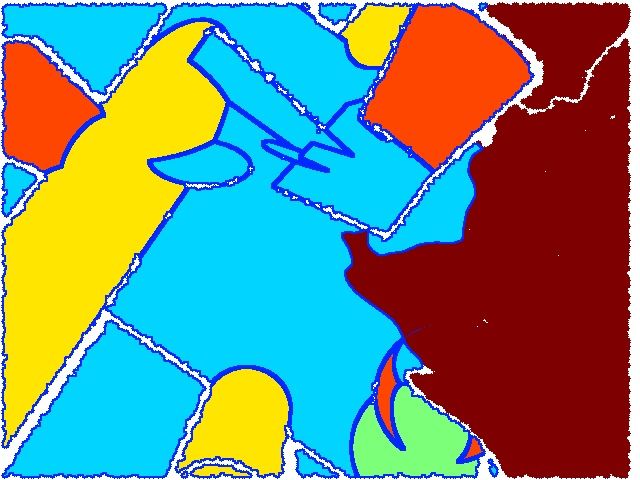}}}\hspace{0em}%
	\frame{\includegraphics[width=.11\textwidth]{{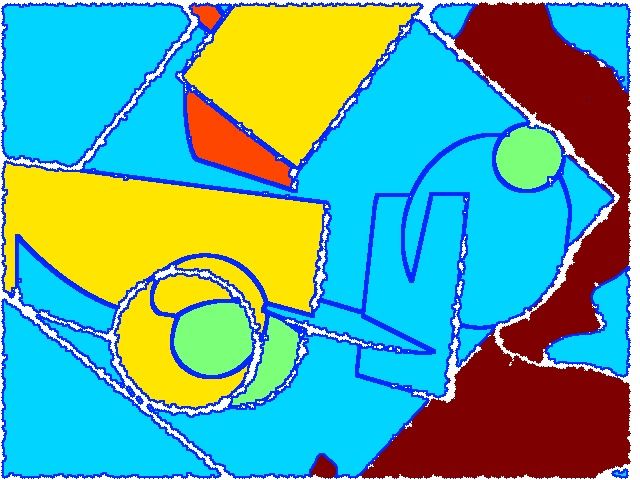}}}\\
	\frame{\includegraphics[width=.11\textwidth]{{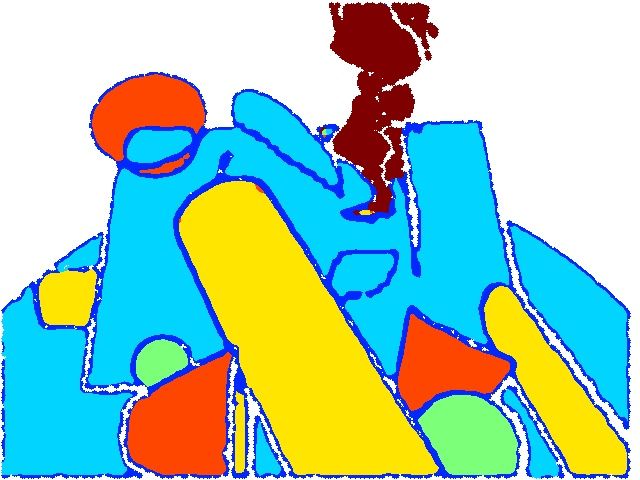}}}\hspace{0em}%
	\frame{\includegraphics[width=.11\textwidth]{{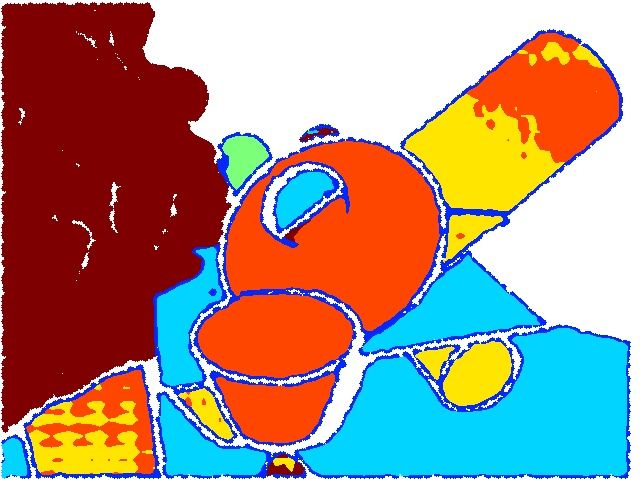}}}\hspace{0em}%
	\frame{\includegraphics[width=.11\textwidth]{{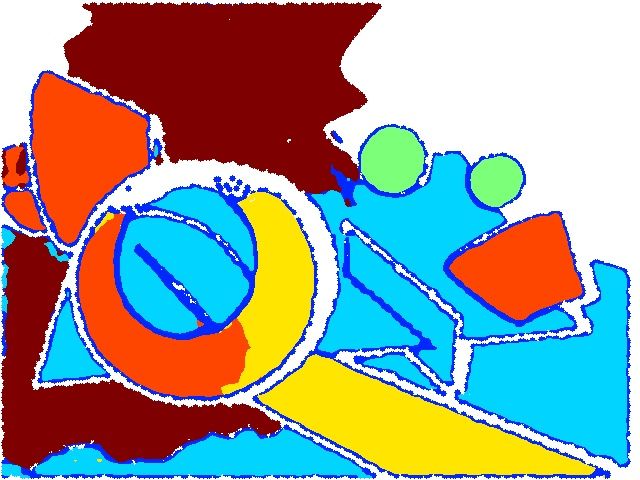}}}\hspace{0em}%
	\frame{\includegraphics[width=.11\textwidth]{{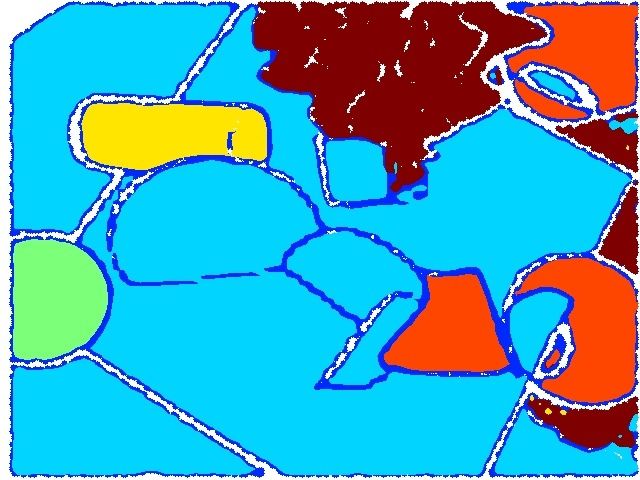}}}\hspace{0em}%
	\frame{\includegraphics[width=.11\textwidth]{{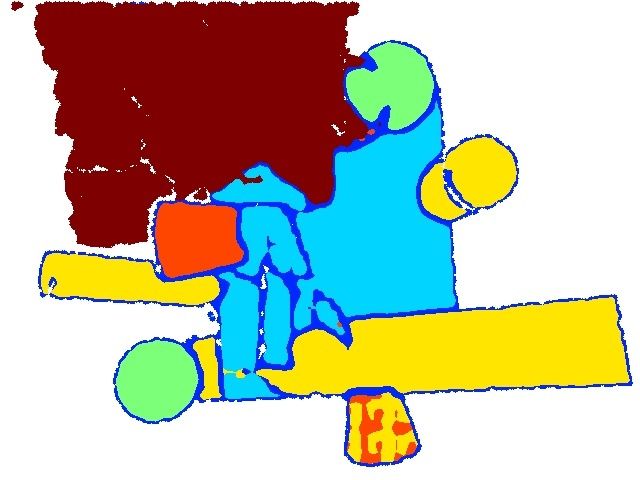}}}\hspace{0em}%
	\frame{\includegraphics[width=.11\textwidth]{{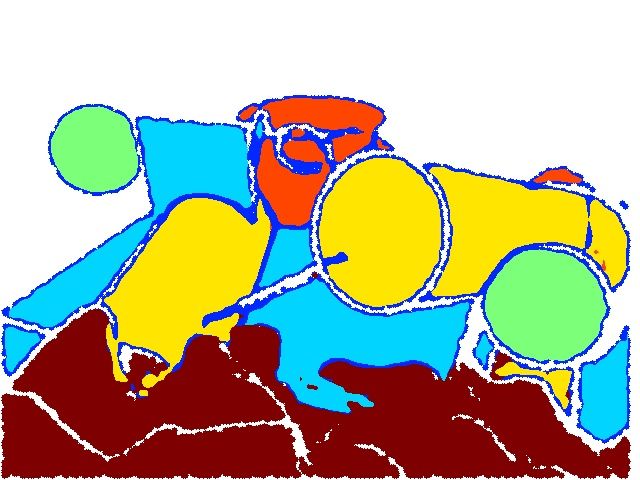}}}\hspace{0em}%
	\frame{\includegraphics[width=.11\textwidth]{{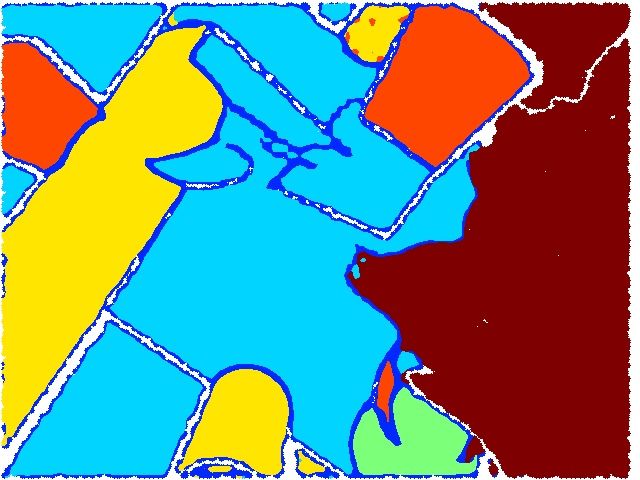}}}\hspace{0em}%
	\frame{\includegraphics[width=.11\textwidth]{{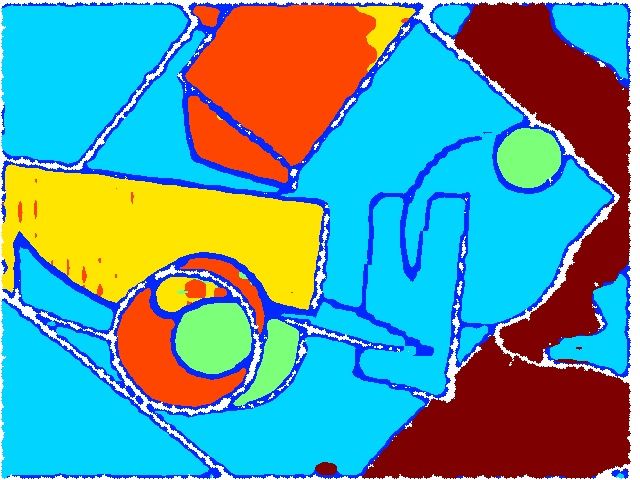}}}\\
	\frame{\includegraphics[width=.11\textwidth]{{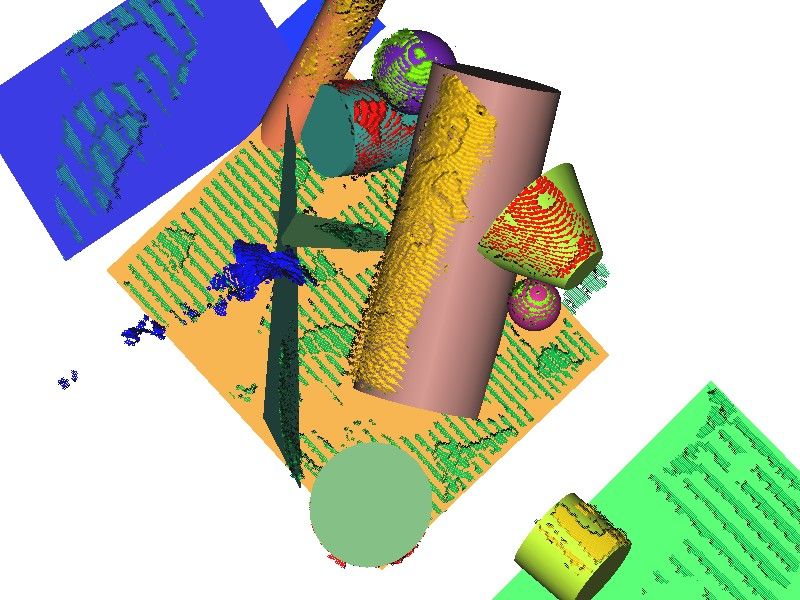}}}\hspace{0em}%
	\frame{\includegraphics[width=.11\textwidth]{{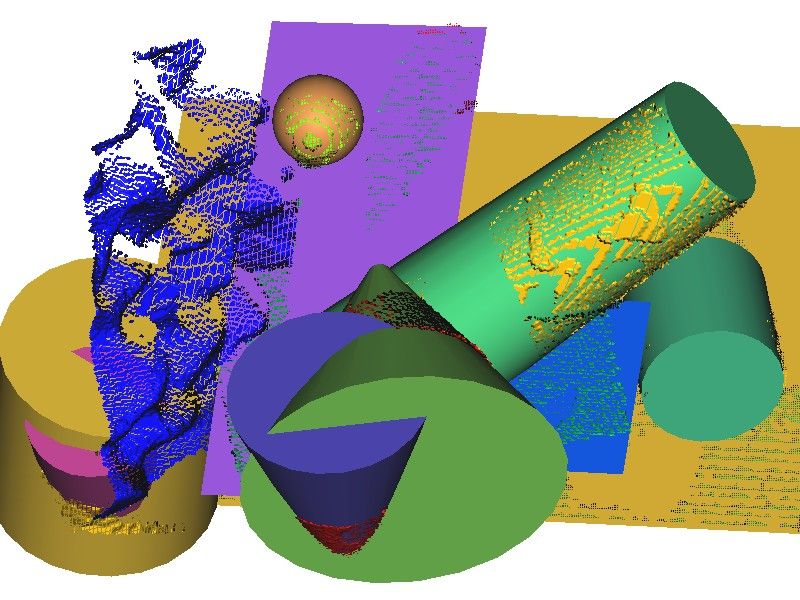}}}\hspace{0em}%
	\frame{\includegraphics[width=.11\textwidth]{{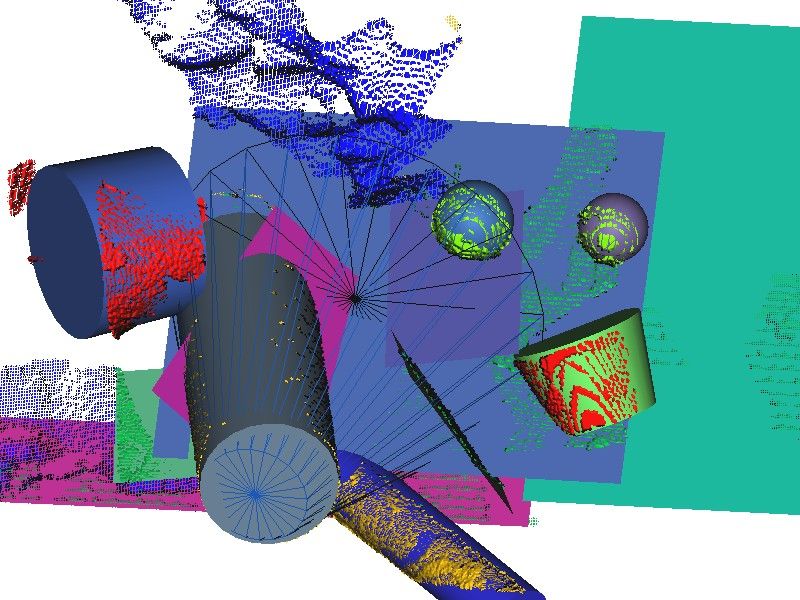}}}\hspace{0em}%
	\frame{\includegraphics[width=.11\textwidth]{{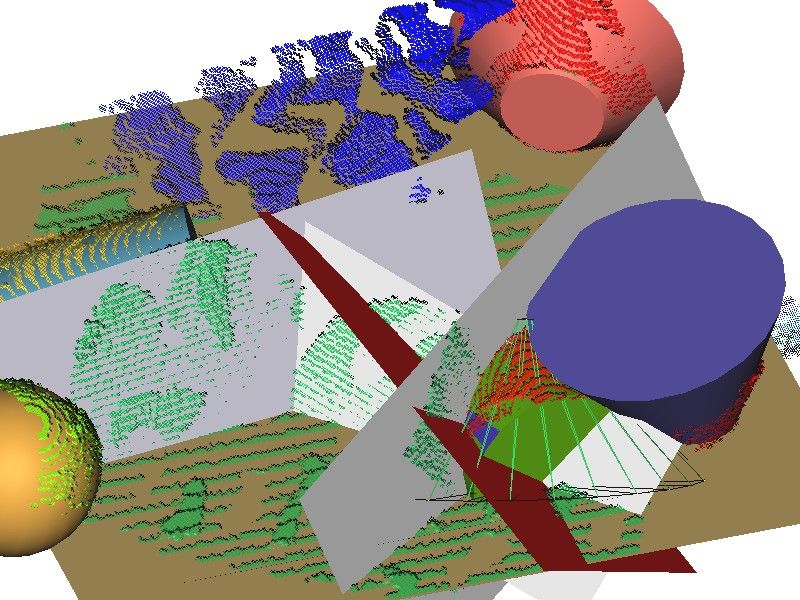}}}\hspace{0em}%
	\frame{\includegraphics[width=.11\textwidth]{{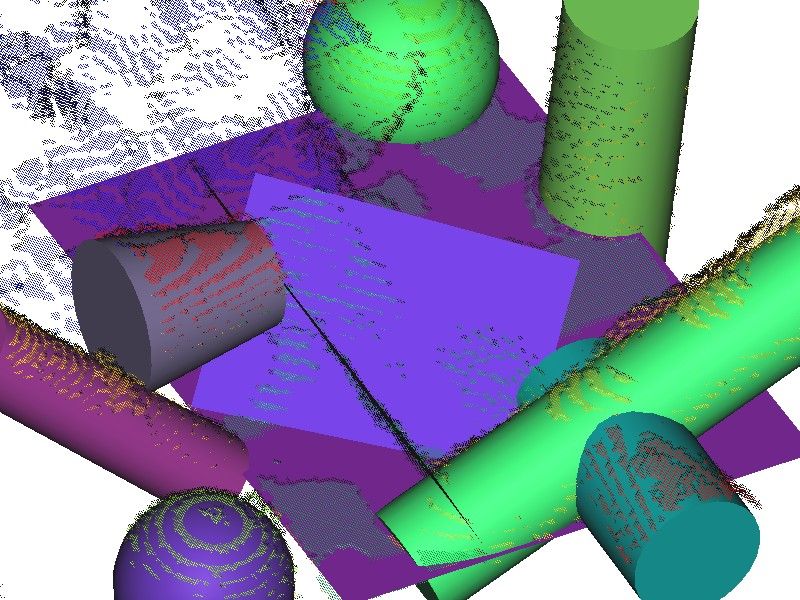}}}\hspace{0em}%
	\frame{\includegraphics[width=.11\textwidth]{{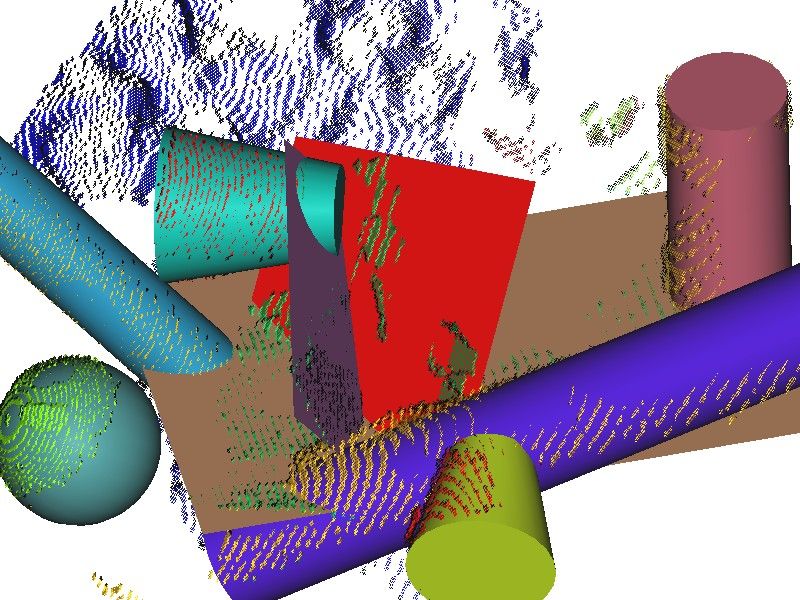}}}\hspace{0em}%
	\frame{\includegraphics[width=.11\textwidth]{{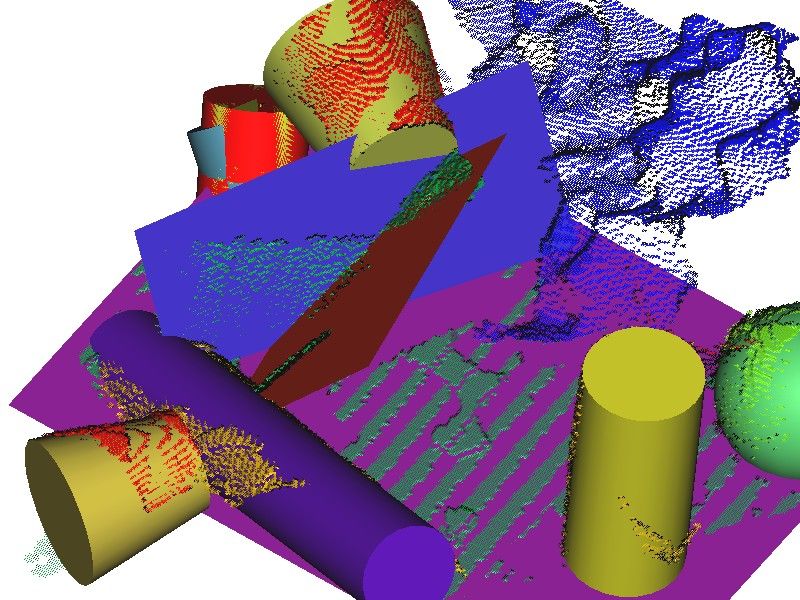}}}\hspace{0em}%
	\frame{\includegraphics[width=.11\textwidth]{{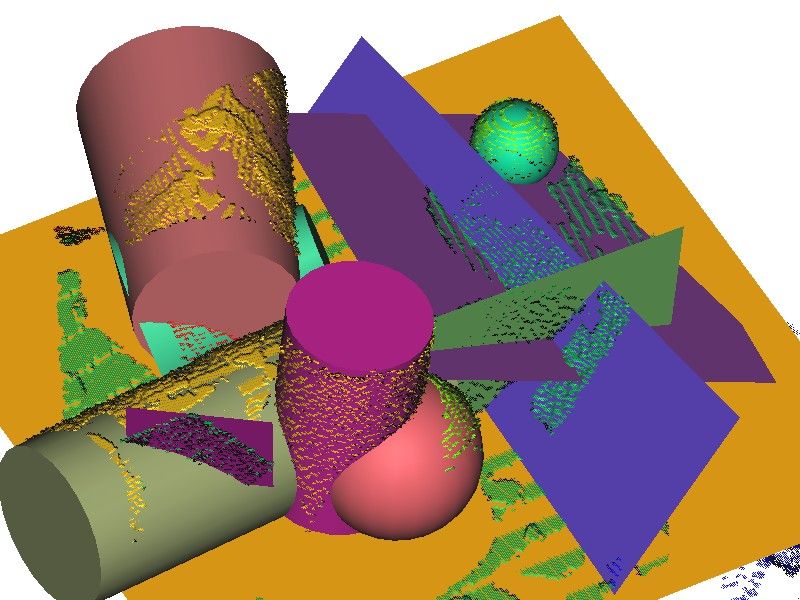}}}
	
	\caption{BAGSFit (\textbf{N5+BAGS}) on simulated test scans. Top: Ground truth labels. Middle: segmentation results. Bottom: fitted primitives (randomly colored) rendered together with real scans.\label{fig:simu}}
\end{figure*}

\textbf{Discussions}. Evaluation results of all 12 networks on the test set of 720 simulated scans are summarized in table~\ref{tab:seg-eval-t}.
\begin{enumerate}
	\item Comparing the \textbf{P/N/PN} rows, we found that normal input turned out to be the best,
	and interestingly outperforming combination of both normal and position.
	This may be caused by the difficulty in normalizing position data for network input.
	
	\item Comparing the \textbf{P/N/PN+MB} rows, we found that the classic multinomial loss leads to better performance mostly than the multi-binomial loss.
	
	\item Comparing the \textbf{N} with \textbf{N+BAGS}, we found that adding additional boundary detection to the segmentation only have very small negative influences to the segmentation performance. This is appealing since we used a single network to perform both segmentation and boundary detection.
	Further comparing the \textbf{N+BAGS} with \textbf{N+BO}, we found that BAGS in fact increases the boundary recall comparing to \textbf{N+BO} that only detects boundaries.
	
	\item Comparing the \textbf{N5} with \textbf{N}, we found that the effect of ignoring background class is inconclusive in terms of significant performance changes,
	which however suggests the benefit of jointly training the background class, as this enables the following steps to focus only on regions seemingly explainable by
	the predefined primitive library.
\end{enumerate}
Just for reference, we tried SVM using neighboring $7\times 7$ or $37\times 37$ normals or principal curvatures for this task,
and the highest pixel-wise accuracy we obtained after many parameter tuning is only 66\%.

\textbf{MaskRCNN}.
We also investigated MaskRCNN~\cite{he2017mask} on this task based on a popular public implementation \url{https://github.com/matterport/Mask_RCNN}, since it is appealing to convert the multi-model problem directly into a single-instance one. In this experiment, the input is a normal map. The network was trained from scratch with the same datasets as BAGS for 100 (instead of 50) epochs with a base lr of 1e-3. With IoU threshold set to 0.5, the mAP of the obtained model was reported to be 26.0\% (36.6\% with IoU threshold of 0.3), which was much lower than its performance on regular RGB object detection tasks. One potential reason might be that the shape of primitives changes significantly compared to natural objects, but the MaskRCNN network has to resize proposed regions of varying aspect ratios into a fixed size. This change of aspect ratio might degrade CNN's performance on this task of a geometric nature. Given this performance, we did not further evaluate its fitting results.

\textbf{Generalizing to Real Data}. Even though we did not tune the simulated scanner's noise model to match our real Kinect scanner,
Figure~\ref{fig:real} shows that the network trained with simulated scans generalizes quite well to real world data.

\subsection{Primitive Fitting Experiments}

For fitting primitives, we used the original efficient RANSAC implementation~\cite{schnabel2007efficient} both as our baseline method (short name \textbf{ERANSAC}) and for our geometric verification.

\textbf{Experiment Details}. We used the following parameters required in~\cite{schnabel2007efficient} for all primitive fitting experiments, tuned on the validation set in effort of maximizing \textbf{ERANSAC} performance:
min number of supporting points per primitive 1000,
max inlier distance 0.03m,
max inlier angle deviation 30 degrees (for counting consensus scores) and 45 degrees (for final inlier set expansion),
overlooking probability 1e-4.
The simulated test set contains
4033 planes,
1256 spheres,
2338 cylinders,
1982 cones,
and in total 9609 primitive instances.

\textbf{Discussions}. Using respective network's segmentation as input to Algorithm~\ref{alg::prim_fit}, the primitive fitting results were evaluated on the simulated test set and summarized in table~\ref{tab:seg-fit-t} together with the \textbf{ERANSAC} baseline. 
\begin{enumerate}
	\item \textbf{ERANSAC} performance is significantly lower than most variants of BAGSFit, in accordance with our qualitative evaluation.
	
	\item \textbf{N5} related experiments receives highest PAP scores, which is reasonable due to the recognition and removal of background classes that greatly reduce the complexity of scenes.
	
	
	\item In terms of average fitting error, \textbf{N}\textbf{+}\textbf{BAGS} $<$ \textbf{N}, \textbf{N5}\textbf{+}\textbf{BAGS} $<$ \textbf{N5}, \textbf{N+MB+BAGS} $<$ \textbf{N+MB}, which strongly supports the benefit of BAGS as mentioned in section~\ref{sec:verify_by_fit}.
	
	\item \textbf{N5+BAGS} gets the lowest fitting error, benefiting from both background and boundary removal before fitting.
\end{enumerate}

\textbf{More results}. Figure~\ref{fig:simu} shows more testing results. The readers may kindly refer to our supplementary video for more result visualizations.

\addtolength{\textheight}{-12cm}   



%

\section*{Acknowledgment}

We thank Yuichi Taguchi, Srikumar Ramalingam, Zhiding Yu, Teng-Yok Lee, Esra Cansizoglu, and Alan Sullivan for their helpful comments.


\bibliographystyle{IEEEtran}
\bibliography{primfit}

\end{document}